\documentclass{article} % For LaTeX2e
\usepackage{iclr2023_conference,times}

\usepackage{amsmath,amsfonts,bm}

\def\eqref#1{equation~\ref{#1}}
\def\1{\bm{1}}

\DeclareMathAlphabet{\mathsfit}{\encodingdefault}{\sfdefault}{m}{sl}
\SetMathAlphabet{\mathsfit}{bold}{\encodingdefault}{\sfdefault}{bx}{n}

\usepackage{hyperref}
\usepackage{url}
\usepackage{float}
\usepackage{algorithm}
\usepackage{algorithmicx, algpseudocode}
\usepackage{graphicx}
\usepackage{bm,multirow}
\usepackage{multicol}
\usepackage{caption}
\usepackage{subcaption}
\usepackage{adjustbox}
\usepackage{bm}

\usepackage{wrapfig}
\def\eqref#1{(\ref{#1})}

\newcommand{\x}{{\boldsymbol x}}
\renewcommand{\d}{{\boldsymbol d}}

\renewcommand{\v}{{\boldsymbol v}}
\newcommand{\e}{{\boldsymbol e}}

\newcommand{\y}{{\boldsymbol y}}
\newcommand{\z}{{\boldsymbol z}}

\newcommand{\Ib}{{\boldsymbol I}}

\newcommand{\Nc}{{\mathcal N}}

\title{%DiffuseIT: Disentangled Image Style Transfer using Diffusion Models
Diffusion-based Image Translation using Disentangled Style and Content Representation
}

\author{{Gihyun Kwon$^1$, Jong Chul Ye$^{2,1}$} \\
Department of Bio and Brain Engineering$^1$, Kim Jaechul Graduate School of AI$^2$,  KAIST\\
\texttt{{cyclomon,jong.ye}@kaist.ac.kr} \\
}

\iclrfinalcopy % Uncomment for camera-ready version, but NOT for submission.
\begin{document}

% \onecolumn[{%
% \renewcommand\twocolumn[1][]{#1}%
\maketitle
\begin{center}
\centering
\includegraphics[width=1.0\linewidth]{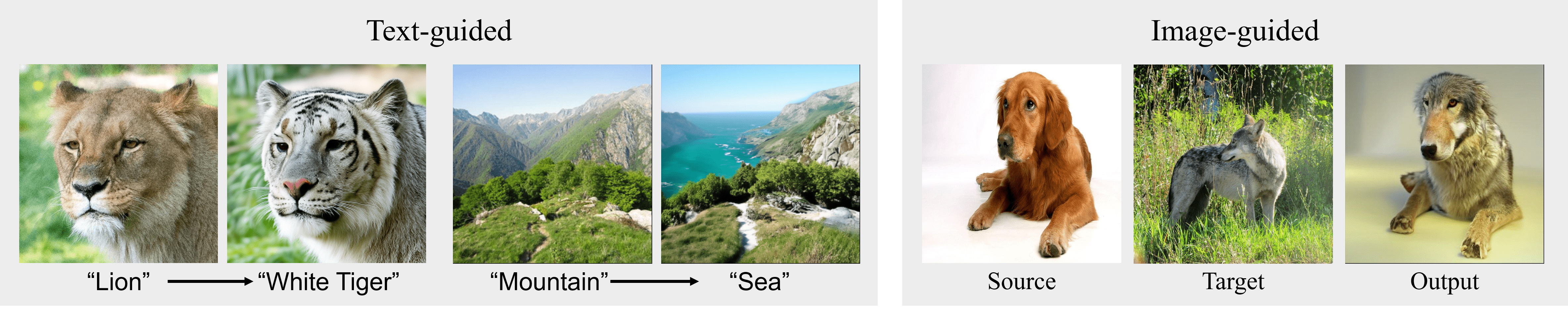}
 \vspace{-0.5cm}
\captionof{figure}{Image translation results by DiffuseIT. Our model can  generate high-quality translation outputs using both text and image conditions. More results can be found in the experiment section. }
\label{fig:first}
\end{center}
% ]

% \maketitle

\begin{abstract}

Diffusion-based image translation guided by  semantic texts   or a single target image   has enabled flexible style transfer which is not limited to the specific domains. 
Unfortunately, due to the stochastic nature of diffusion models, 
%unless the diffusion model is trained with matched target data,
it is often  difficult to maintain the original content of the image  during the reverse diffusion.
%unless conditional diffusion models trained with matched targets are used. 
To address this, here
we present a novel diffusion-based unsupervised image translation method, dubbed as {\em DiffuseIT}, using disentangled style and content representation.
 Specifically, inspired by the  slicing Vision Transformer  \citep{splice},
% can convert the semantic appearance of a given image into target domain while maintaining the structure of input image, our method
  we extract intermediate keys of multihead self attention layer  from ViT model and used them as the content preservation loss. % More specifically, to preserve the structure information we use the contrastive loss between intermediate keys of the input image and the estimated denoised output during the reverse diffusion sampling. 
  Then, an image guided style transfer is performed by matching the [CLS] classification token from the denoised samples and target image, whereas additional CLIP loss is used for the text-driven style transfer.
% 
% preserve the content structures between input and output using .
%% 
%Image Style Transfer using Diffusion Models
%
% a new method called diffusion-based image translation , which enables text-guided image translation using Denoising Diffusion Probabilistic Model and text-image embedding model CLIP. First, our method proposes a  content preservation loss with leveraging the pre-trained Vision Transformer model to preserve the content structures between input and output. Second,
  To further accelerate the semantic change during the reverse  diffusion, we also propose a novel semantic divergence loss and resampling strategy. 
 %Third, we propose a resampling strategy to obtain the good initial point of the reverse step of DDPM. Furthermore, we show that our model can be easily adapted to image-guided image translation framework. 
 Our experimental results show that the proposed method outperforms state-of-the-art baseline models in both text-guided and image-guided translation tasks. 
\end{abstract}

\section{Introduction}

Image translation is a task in which the model receives an input image and converts it into a target domain. Early  image translation approaches \citep{cyclegan,cut,pix2pix} were mainly designed for single domain translation, but soon extended to multi-domain translation  \citep{stargan,collagan}. %Recently, multi-modal image translation methods that can generate diverse images (\cite{starganv2,tunit}) simultaneously with domain transformation have also been proposed and showed high performance. 
As these methods demand large training set for each domain, image translation approaches using only a single image pairs have been studied,
which include the one-to-one image translation using multiscale training \citep{tuigan}, or patch matching strategy \citep{gpen,strotss}. Most recently, Splicing ViT \citep{splice} exploits a pre-trained DINO ViT \citep{dino} to convert the semantic appearance of a given image into a target domain while maintaining the structure of input image. 

On the other hand,  by employing the recent text-to-image embedding model such as CLIP \citep{clip_rad},
%which show outstanding performance, 
several approaches have attempted %proposed to combine the model with image generation models in order 
to generate  images conditioned on text prompts \citep{styleclip,nada,vqclip,flexit}.
%In previous studies, most text-guided image generation methods (\cite{styleclip,nada,vqclip,flexit}) used 
As these methods rely on Generative Adversarial Networks (GAN) as a backbone generative model,
 the semantic changes are not often properly controlled  %during the GAN inversion
  when applied to an out-of-data (OOD)
image generation.
% as the model is limited to specific data domain, 
 
 %and an GAN inversion process is additionally required for real image editing. 
Recently, score-based generative models \citep{ddpm,scoresde,iddpm} have demonstrated state-of-the-art performance in text-conditioned image generation  \citep{dalle2,imagen,cgd,blended}. However, when it comes to the image translation scenario in which multiple conditions (e.g. input image, text condition) are given  to the score based model, disentangling and separately controlling the components  still remains as an open problem. 

In fact, one of  the most important open questions  in image translation by diffusion models is to transform only the semantic information (or style) while maintaining the structure information (or content) of the input image. 
Although this could not be an issue with the conditional diffusion models trained with matched input and target domain images \citep{saharia2022palette},
such training is impractical in many image translation tasks (e.g. summer-to-winter, horse-to-zebra translation).
On the other hand, existing  methods using unconditional diffusion models often fail to preserve content information due to the entanglement problems in which semantic and content change at the same time \citep{blended,cgd}.
% as they do not consider the structure of input image .
DiffusionCLIP \citep{diffusionclip} tried to address this problem using denoising diffusion implicit models (DDIM) \citep{song2020denoising}
and  pixel-wise loss, but
the score function needs to be fine-tuned for a novel target domain, which is computationally expensive.
 
% Although these methods can manipulate the images with text conditions,  
% In order to disentangle the style and content components and separately control each, we introduced a loss function using pre-trained Vision Transformer(ViT) model(\cite{vit}). Based on the recent idea (\cite{splice}), we extract intermediate keys of multihead self attention layer and [CLS] tokens the last layer from ViT model and used them as our content and style regularization, respectively. 
In order to control the diffusion process in such a way that it produces the output that simultaneously retain the content of the input image and follow the semantics of the target text or image,
here we introduce a loss function using a pre-trained Vision Transformer (ViT)  \citep{vit}. Specifically, inspired by the recent idea \citep{splice}, we extract intermediate keys of multihead self attention layer and [CLS] classification tokens of the last layer from the DINO ViT model and used them as our content and style regularization, respectively. 
More specifically,  to preserve the structural information, we  use  the similarity and contrastive  loss between intermediate keys of the input and  denoised
image during the sampling. 
Then, an image guided style transfer is performed by matching the [CLS] token between the denoised sample and the target domain,
whereas additional CLIP loss is used for the text-driven style transfer.
To further improve the sampling speed, 
 we propose a novel semantic divergence loss and resampling strategy.
 
% to accelerate the semantic change of output. Third, to further improve the generation process, we propose resampling strategy in order to guide the initial point of diffusion process to better point. With the above approaches, we can successfully guide the diffusion model to generate outputs with semantic change. In addition, we can easily adapt our framework to image-guided style translation task.
 
Extensive experimental results including Fig.~\ref{fig:first}
confirmed that our method provide state-of-the-art performance in both text- and image- guided style transfer  tasks quantitatively and qualitatively.
To our best knowledge, this is the first unconditional diffusion model-based image translation method that allows both text- and image- guided style transfer without altering input
image content.

\section{Related Work}
%\textbf{Image Translation. }
%Image translation is a task in which the model receives an input image and converts it into target domain. Initial approaches of image translation(\cite{cyclegan,cut,pix2pix}) are mainly for single domain, but it extended to models for multi-domain translation later(\cite{stargan,collagan}). Recently, multi-modal image translation methods that can generate diverse images(\cite{starganv2,tunit}) simultaneously with domain transformation have also been proposed and showed high performance. At the same time, in contrast to the above methods which use large training set, novel attempts of image translation with only a single image pairs have been recently proposed. There are methods for one-to-one image translation using multiscale training(\cite{tuigan}), or patch matching strategy(\cite{gpen,strotss}). Most recently, Splicing Vit(\cite{splice}) exploited pre-trained DINO ViT(\cite{dino}) to convert the semantic appearance of given image into target domain while maintaining the structure of input image. 
%

\textbf{Text-guided image synthesis.}
Thanks to the  outstanding performance of text-to-image alignment in the feature space, CLIP has been widely used in various text-related computer vision tasks including object generation \citep{fusedream,clipnerf}, style transfer \citep{clipstyler,ldast}, object segmentation \citep{clipseg,cris}, etc. Several recent approaches also demonstrated state-of-the-art performance in text-guided image manipulation task by combining the CLIP with image generation models. Previous approaches leverage pre-trained StyleGAN \citep{stylegan2} for image manipulation with a text condition \citep{styleclip,nada,hairclip}. However, StyleGAN-based methods cannot be used in arbitrary natural images since it is restricted to the pre-trained data domain. Pre-trained VQGAN \citep{vqgan} was proposed
for better generalization capability in the image manipulation, but it often suffers from poor image quality due to limited power of the backbone model. 
 
With the advance of score-based generative models such as Denoising Diffusion Probabilistic Model (DDPM) \citep{ddpm}, several methods \citep{dalle2,imagen} tried to generate photo-realistic image samples with given text conditions. However, these approaches are not adequate for image translation framework
 as the text condition and input image are not usually disentangled. %, therefore both components cannot be separately controlled. 
 Although DiffusionCLIP \citep{diffusionclip} partially solves the problem  using DDIM sampling and pixelwise regularization during the reverse diffusion, it has major disadvantage in that it requires fine-tuning process of score models. 
 %Our approach, DiffuseIT propose novel methods to guide pre-trained DDPM model so that we can solely change the semantic style of give image without changing the overall structure.

\textbf{Single-shot Image Translation.}
In image translation using single target image, early models mainly focused on image style transfer \citep{nst,adain,sanet,wct2}. Afterwards, methods using StyleGAN adaptation \citep{fsga,mind,oneshotclip,jojogan} showed great performance, but there are limitations as the models are domain-specific (e.g. human faces). In order to overcome this, methods for converting unseen image into a semantic of target \citep{tuigan,strotss,gpen} have been proposed, but these methods often suffer from degraded image quality. Recently, Splicing ViT \citep{splice} successfully exploited pre-trained DINO ViT\citep{dino} to convert the semantic appearance of given image into target domain while preserving the structure of input.

 % \add{JUST FOCUS ON SINGLE-SHOT IMAGE TRANSLATION STARTING FROM ADAIN. Image translation is a task in which the model receives an input image and converts it into target domain.  So far, various models\citep{cyclegan,cut,pix2pix,stargan,collagan,starganv2,tunit} have shown high performance by training on large datasets. 
 % % Initial approaches of image translation(\cite{cyclegan,cut,pix2pix}) are mainly for single domain, but it extended to models for multi-domain translation later(\cite{stargan,collagan}). Recently, multi-modal image translation methods that can generate diverse images(\cite{starganv2,tunit}) simultaneously with domain transformation have also been proposed and showed high performance. 
 % At the same time, in contrast to the above methods, novel attempts of image translation with only a single image pairs have been recently proposed. There are methods for one-to-one image translation using multiscale training\citep{tuigan}, or patch matching\citep{gpen,strotss}. Most recently, Splicing Vit\citep{splice} exploited pre-trained DINO ViT\citep{dino} to convert the semantic appearance of given image into target domain while maintaining the structure of input. }
 % % \add{ADD REVIEW OF ONE-shot image translation}
 
\section{Proposed Method}

\subsection{DDPM Sampling with Manifold Constraint}

In DDPMs~\citep{ddpm}, starting from a clean image $\x_0 \sim q(\x_0)$,  a forward diffusion process $q(\x_t|\x_{t-1})$ is described as a Markov chain 
that gradually adds Gaussian noise at every time steps $t$: %. The following forward diffusion process can be expressed as 
\begin{align}
    q(\x_{T}|\x_0):=\prod_{t=1}^{T}q(\x_t|\x_{t-1}),\quad \mbox{where}\quad
    q(\x_t|\x_{t-1}):=\Nc(\x_t;\sqrt{1-\beta_t}\x_{t-1},\beta_t\Ib),
\end{align}
where $\{\beta\}_{t=0}^{T}$ is a variance schedule.
By denoting $\alpha_t:=1-\beta_t$ and $\bar{\alpha_t}:=\prod_{s=1}^t\alpha_s$, the
forward diffused sample at $t$, i.e. 
$\bm x_t$, can be sampled in one step as:
%we can sample from $q(\x_t|\x_0)$ in a closed form:
\begin{align}
    \label{eq:ddpm}
    \x_t=\sqrt{\bar{\alpha}_t}\x_0+\sqrt{1-\bar{\alpha}_t}\boldsymbol{\epsilon},\quad \mbox{where}\quad \boldsymbol{\epsilon}\sim\Nc(\textbf{0},\Ib).
\end{align}
As the reverse of the forward step $q(\x_{t-1}|\x_t)$ is intractable, DDPM learns to maximize the variational
lowerbound through a
 parameterized Gaussian transitions $p_\theta(\x_{t-1}|\x_t)$ with the parameter $\theta$. 
Accordingly, the reverse process is approximated as Markov chain with learned mean and fixed variance, starting from $p(\x_T)=\Nc(\x_T;\textbf{0},\boldsymbol{I})$:
\begin{align}
    p_\theta(\x_{0:T}):=p_\theta(\x_T)\prod_{t=1}^{T}p_\theta(\x_{t-1}|\x_{t}),\quad \mbox{where}\quad
    p_\theta(\x_{t-1}|\x_{t}):=\Nc(\x_{t-1};{\boldsymbol\mu}_\theta(\x_t,t),\sigma_t^2\Ib).
\end{align}
where
\begin{align}\label{eq:mu}
    {\boldsymbol\mu}_\theta(\x_t,t):=\frac{1}{\sqrt{\alpha}_t}\Big{(}\x_t-\frac{1-\alpha_t}{\sqrt{1-\bar{\alpha}_t}}\boldsymbol{\epsilon}_\theta(\x_t,t)\Big{)},
    %\quad     \boldsymbol{s}_\theta(\x_t,t)=-\frac{1}{\sqrt{1-\bar{\alpha}_t}}\z_\theta(\x_t,t)
\end{align}
Here, $\boldsymbol{\epsilon}_\theta(\x_t,t)$ is the diffusion model trained 
%The training cost function is performed
 by optimizing the objective:
\begin{align}
    \label{eq:objective}
    \min_{\theta} L(\theta), \quad\mbox{where}\quad L(\theta):=\mathbb{E}_{t,\x_0,\boldsymbol{\epsilon}}\Big{[}\|\boldsymbol{\epsilon}-\boldsymbol{\epsilon}_\theta(\sqrt{\bar{\alpha}_t}\x_0+\sqrt{1-\bar{\alpha}_t}\boldsymbol{\epsilon},t)\|^2\Big{]}.
\end{align}
After the optimization, by plugging learned score function into the generative (or reverse) diffusion process, one can simply sample  from $p_\theta(\x_{t-1}|\x_t)$ by
\begin{align}\label{eq:reverse}
    \x_{t-1}=   {\boldsymbol\mu}_\theta(\x_t,t)+\sigma_t\boldsymbol{\epsilon}=\frac{1}{\sqrt{\alpha_t}}\Big{(}\x_t-\frac{1-\alpha_t}{\sqrt{1-\bar{\alpha}_t}}\boldsymbol{\epsilon}_\theta(\x_t,t)\Big{)}+\sigma_t\boldsymbol{\epsilon}
\end{align}

In image translation using {\em conditional} diffusion models \citep{saharia2022palette,unitddpm}, the diffusion model $\boldsymbol{\epsilon}_\theta$ in
\eqref{eq:objective} and \eqref{eq:reverse} should be replaced with
$\boldsymbol{\epsilon}_\theta(\y, \sqrt{\bar{\alpha}_t}\x_0+\sqrt{1-\bar{\alpha}_t}\boldsymbol{\epsilon},t)$ where $\y$ denotes the matched target image.
Accordingly, the sample generation is tightly controlled by the matched target in a supervised manner, so that the image content change rarely happen.
Unfortunately, the requirement  of the {\em matched} targets for the training makes this approach impractical.

 To address this, \cite{diffusionbeat} proposed classifier-guided image translation using
  the unconditional diffusion model training as in \eqref{eq:objective} and a pre-trained classifier $p_{\phi}(\y|\x_t)$. 
  Specifically, ${\boldsymbol\mu}_\theta(\x_t,t)$ in \eqref{eq:mu} and \eqref{eq:reverse} are supplemented with the gradient of the classifier, i.e.
  ${\bm \hat\mu}_\theta(\x_t,t):={\boldsymbol\mu}_\theta(\x_t,t)+\sigma_t \nabla_{\x_t} \log p_{\phi}(\y|\x_t)$.
However, most of the classifiers, which should be separately trained, are not usually
  sufficient to control the content of the samples from the reverse diffusion process.

Inspired by the recent manifold constrained gradient (MCG)  for inverse problems \citep{mcg}, here we formulate
our content and style guidance problem as an inverse problem, which can be solved by minimizing the following total cost function with respect to the sample $\x$:
\begin{align}\label{eq:total}
\ell_{total}(\x;\x_{trg},\x_{src}),\quad \mbox{or}\quad \ell_{total}(\x;\d_{trg},\x_{src},\d_{src})
\end{align}
where 
$\x_{src}$ and $\x_{trg}$ refer to the
source and target images, respectively;
and $\d_{src}$ and $\d_{trg}$ refer to the
source and target text, respectively.
 In our paper, the first form of the total loss in \eqref{eq:total} is used for image-guided translation,
 where the second form is for the text-guided translation. 
Then, the sampling from the reverse diffusion with MCG is given by
\begin{align}\label{eq:MCG}
    \x'_{t-1}&=\frac{1}{\sqrt{\alpha_t}}\Big{(}\x_t-\frac{1-\alpha_t}{\sqrt{1-\bar{\alpha}_t}}\boldsymbol{\epsilon}_\theta(\x_t,t)\Big{)} +\sigma_t\boldsymbol{\epsilon}\\
    \x_{t-1}&= \x'_{t-1}
    - \nabla_{\x_t} \ell_{total}(\hat\x_0(\x_{t}))
    \end{align}
where $\hat\x_0(\x_{t})$ refers to the
estimated clean image  from the sample $\x_t$  using the Tweedie's formula \citep{noise2score}:
\begin{align} \label{eq:4}
\hat{\x}_0(\x_t) := \frac{\x_t}{\sqrt{\Bar{\alpha}_t}} - \frac{\sqrt{1-\Bar{\alpha}_t}}{\sqrt{\Bar{\alpha}_t}}{\bm \epsilon}_\theta(\x_t,t).
\end{align}
%where $\hat{x}^t_0$ indicates estimated clean image $x_0$ from time $t$. 

%{However, since our goal is to guide the diffusion process with ViT and CLIP models pre-trained on clean images, we cannot directly use the above classifier-guidance framework. With estimated clean image $\hat{x}_0$, We can now calculate the loss function $\mathcal{L}$ and use the gradient as guidance for diffusion process such as: $x_{t-1}\sim\mathcal{N}(\mu+\Sigma\nabla_{\hat{x}^t_0} \mathcal{L},\Sigma)$. The loss functions proposed later will be based on this denoising framework of using $\hat{x}_0$.} 

In the following, we describe how the total loss $\ell_{total}$ is defined. For brevity, we notate $\hat{\x}_0(\x_t)$ as $\x$ in the following sections.

% we can calculate the loss function with pre-trained CLIP model.

% proposed in the recent work(\cite{iddpm}), instead of fixing the variance(\cite{ddpm}), we use learnable variance as it showed better performance with reducing the sampling steps. 

\begin{figure}[t!]
    \includegraphics[width=0.95\linewidth]{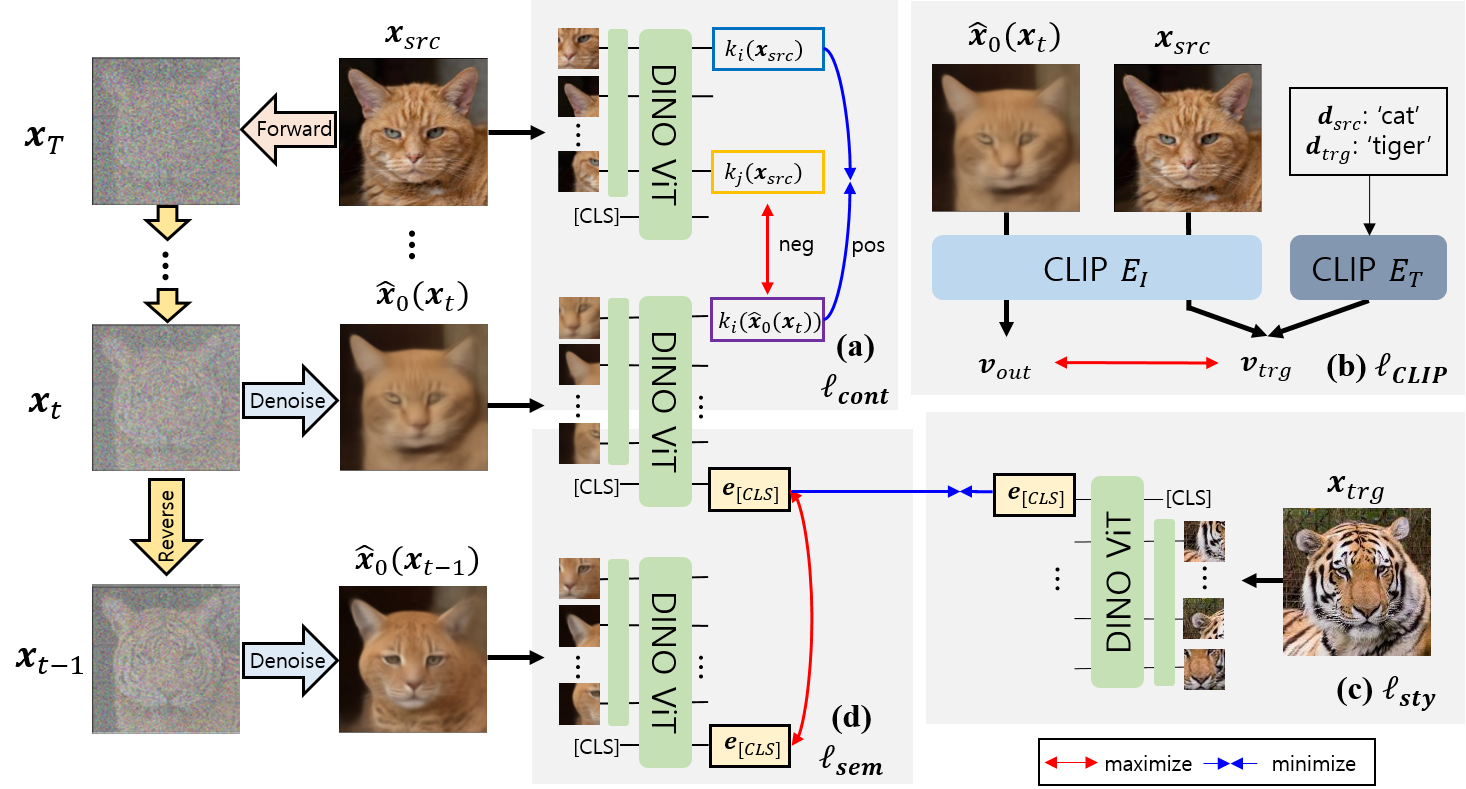}
    \vspace{-0.3cm}
    \caption{Given the input image $\x_{src}$, we guide the reverse diffusion process $\{\x_t\}_{t=T}^0$ using various losses. (a) 
    $\ell_{cont}$:   the structural similarity loss between input and outputs in terms of contrastive loss
    between extracted keys from ViT.
     (b) $\ell_{CLIP}$: relative distance to the target text $\d_{trg}$ in CLIP space in terms of $\x_{src}$ and $\d_{src}$. (c) $\ell_{sty}$: the [CLS] token distances between the outputs and target $\x_{trg}$. (d) $\ell_{sem}$:  dissimilarity between the [CLS] token from the  present and past denoised samples.}
    \label{fig:fig_method}
    \end{figure}

\subsection{Structure Loss} \label{sec:structure}
% Although the above loss can guide the generation process with text condition, the loss is not enough for our image translation framework. 
As previously mentioned, the main objective of image translation is maintaining the content structure between output and the input image, while guiding the output to follow semantic of target condition. Existing methods \citep{flexit,diffusionclip} use pixel-wise  loss or the perceptual loss for the content preservation. However, the pixel space does not explicitly discriminate content and semantic components:  too strong pixel loss hinders the semantic change of output, whereas weak pixel loss alters  the structural component along with semantic changes. To address the problem, we need to separately process the semantic and structure information of the image. 

Recently, \citep{splice} demonstrated successful disentanglement of both components using a pre-trained DINO ViT \citep{dino}. They showed that in ViT, the keys $k^l$ of multi-head self attention (MSA) layer contain structure information, and [CLS] token of last layer contains the semantic information. With above features, they proposed a loss for maintaining structure between input and network output with matching the self similarity matrix $S^l$ of the keys, which can
be represented in the following form for our problem: %the loss could be described as:
\begin{align}
\ell_{ssim}(\x_{src},\x) = \|S^l(\x_{src})-S^l(\x)\|_F, \quad \mbox{where} \quad \left[S^l(\x)\right]_{i,j} = \cos(k_i^l(\x),k_j^l(\x)), 
\end{align}
where $k_i^l(\x)$ and $k_j^l(\x)$ indicate $i,j$th key in the $l$-th MSA layer extracted from ViT with image $\x$. % $x_{src}$ and $\hat{x}_0^t$ are input and denoised output images at time $t$.
The self-similarity loss can maintain the content information between input and output, but we found that only using this loss results in a weak regularization in our DDPM framework. Since the key $k_i$ contains the spatial information corresponding the $i$-th patch location,
we use additional regularization with contrastive learning as shown in Fig.~\ref{fig:fig_method}(a), inspired by the idea of using both of relation consistency and contrastive learning\citep{HnegSRC}. 
Specifically, leveraging the idea of patch contrastive loss \citep{cut}, we define the infoNCE loss using  the DINO ViT keys:
\begin{align}
 \ell_{cont}(\x_{src},\x) = -\sum_{i}\log\left(\frac{\exp(\mathrm{sim}({k}^l_i(\x), {k}^l_i(\x_{src}))/\tau)}{ \exp( \mathrm{sim}({k}^l_i(\x), {k}^l_i(\x_{src}))/\tau + \sum_{j\neq i} \exp(\mathrm{sim}({k}^l_i(\x), {k}^l_j(\x_{src}))/\tau)}\right),
\end{align}
where $\tau$ is temperature, and $\mathrm{sim}(\cdot,\cdot)$ represents the normalized cosine similarity. With this loss, we regularize the key of same positions to have closer distance, while maximizing the distances between the keys at different positions.

\subsection{Style Loss}
\paragraph{CLIP Loss for Text-guided Image Translation}
Based on the previous work of \citep{diffusionbeat}, CLIP-guided diffusion \citep{cgd} proposed to guide the reverse diffusion using pre-trained CLIP model
using the following loss function:
% Based on the previous work of Dhariwal and Nichol(\cite{diffusionbeat}), which guided the reverse process of diffusion model using a classifier trained on noisy images, CLIP-guided diffusion(\cite{cgd}) proposed to guide the reverse process using pre-trained CLIP model. Unlike classifier guidance, since the CLIP model is trained on clean images, we have to estimate the denoised clean image at intermediate step of reverse process to guide the generation process.
% In order to estimate the clean image from intermediate steps, we can obtain estimated clean image $\hat{x}_0$ from intermediate output $x_t$  with using equation \ref{eq:2}:
% \begin{align} \label{eq:4}
% \hat{x}^t_0 = \frac{x_t}{\sqrt{\Bar{\alpha}_t}} = (x_t - \frac{\sqrt{1-\Bar{\alpha}_t}}{\sqrt{\Bar{\alpha}_t}}\epsilon_\theta(x_t,t)),
% \end{align}
% where $\hat{x}^t_0$ indicates estimated clean image $x_0$ from time $t$. With estimated clean image $\hat{x}_0$, we can calculate the loss function with pre-trained CLIP model. 
%In the vanilla version of CLIP-guided diffusion, the loss function is simply defined as:
\begin{align}  \label{eq:5}
\ell_{CLIP}(\d_{trg}, \x) := -\mathrm{sim}\left(E_{T}(\d_{trg}),E_{I}(\x)\right),
\end{align}
where $\d_{trg}$ is the target text prompt, and $E_{I}, E_{T}$ refer to the image and text encoder of CLIP, respectively. 
Although this loss can give text-guidance to diffusion model, the results often suffer from poor image quality. 

Instead,  we propose to use input-aware directional CLIP loss (\cite{nada}) which matches the CLIP embedding of the output image to the target vector 
in terms of $\d_{trg}$, $\d_{src}$, and $\x_{src}$. More specifically, our CLIP-based semantic loss is described as  (see also Fig.~\ref{fig:fig_method}(b)):
\begin{align}
\ell_{CLIP}(\x; \d_{trg}, \x_{src},\d_{src}) := - \mathrm{sim}(\v_{trg},\v_{src})
\end{align}
where
\begin{align}
%- \frac{v_{trg} \cdot v_{out}}{|v_{trg}| \cdot |v_{out}|},
%\end{align}
%\begin{gather}
\v_{trg}:= E_T(\d_{trg}) + \lambda_i E_I(\x_{src}) - \lambda_s E_T(\d_{src}),&\quad \v_{src}:=  E_I(\mathrm{aug}(\x)) 
\end{align}
where 
$\mathrm{aug(\cdot)}$ denotes the augmentation for preventing adversarial artifacts from CLIP.
Here, we simultaneously remove the source domain information $- \lambda_s E_T(\d_{src})$ and reflect the source image information to output $+ \lambda_i E_I(\x_{src})$ according to the values of $\lambda_s$ and $\lambda_i$. Therefore it is possible to obtain stable outputs compared to using the conventional loss. 

Furthermore, in contrast to the existing methods using only single pre-trained CLIP model (e.g. ViT/B-32), we improve the text-image embedding performance by using the recently proposed CLIP model ensemble method (\cite{flexit}). Specifically, instead of using a single embedding, we concatenate the multiple embedding vectors from {multiple pre-trained CLIP models} and used the it as our final embedding.

\paragraph{Semantic Style Loss for Image-guided Image Translation}
In the case of image-guide translation, we propose to use {[CLS] token} of ViT as our style guidance. As explained in the previous part \ref{sec:structure}, the [CLS] token contains the semantic style information of the image. Therefore, we can guide the diffusion process to match the semantic of the samples to that of target image by minimizing the [CLS] token distances as shown in Fig. \ref{fig:fig_method}(c). Also, we found that using only [CLS] tokens often results in misaligned color values. To prevent this, we guide the output to follow the overall color statistic of target image with weak MSE loss between the images. Therefore, our loss function is described as follows:
\begin{align}
\ell_{sty}(\x_{trg},\x) = ||\e_{[CLS]}^L(\x_{trg})-\e_{[CLS]}^L({\x})||_2 + \lambda_{mse}||\x_{trg}-{\x}||_2.
\end{align}
where $\e_{[CLS]}^L$ denotes the last layer  [CLS] token.

\subsection{Acceleration Strategy}

\paragraph{Semantic Divergence Loss}\label{sec:sem}
% As explained in the previous part, the [CLS] token of ViT contains the overall semantic information of the image. In the previous method\cite{splicing}, the output was guided to follow the reference semantic by matching the [CLS] token distance between the output image and the reference target image. 
With the proposed loss functions, we can achieve text- or image-guided image translation outputs. However, we empirically observed that the generation process requires large steps to reach the the desired output. To solve the problem, we propose a simple approach to accelerate the diffusion process. As explained before, the [CLS] token of ViT contains the overall semantic information of the image. Since our purpose is to make the semantic information as different from the original as possible while maintaining the structure, we conjecture that we can achieve our desired purpose by maximizing the distance between the [CLS] tokens of the previous step and the current output during the generation process as described in Fig.~\ref{fig:fig_method}(d). Therefore,  our loss function at time $t$ is given by
\begin{align}
 \ell_{sem}(\x_t;\x_{t+1}) = -||\e_{[CLS]}^L(\hat{\x}_0(\x_t))-\e_{[CLS]}^L(\hat{\x}_0(\x_{t+1}))||_2,
\end{align}
%where $e_{[CLS]}^L(I)$ is [CLS] token extracted from the deepest layer $L$ of ViT. In our framework, 
Specifically, we maximize the distance between the denoised output of the present time  and the previous time, so that next step sample has different semantic from the previous step.  One could think of alternatives to maximize pixel-wise or perceptual distance, but we have experimentally found that in these cases, the content structure is greatly harmed. In contrast, our proposed loss has advantages in terms of image quality because it can control only the semantic appearance.

\paragraph{Resampling Strategy}
As shown in CCDF acceleration strategy \citep{ccdf}, a better initialization leads to an accelerated reverse diffusion for inverse problem.
Empirically, in our image translation problem we also find that finding the good starting point at time step $T$ for the reverse diffusion affects the overall image quality.
%
%
%
% Although giving guidance to the reverse step with the proposed loss shows good results, the output quality is not stable since our framework of DDPM has stochastic randomness. We found that this is greatly affected by the initial forward sampling step $q(x_t|x_0)$.
% 
%  For stable and faster reverse process, we propose to use the resampling strategy based on the idea from Repaint(\cite{repaint}). Different from Repaint which matches the intermediate noisy images with noisy ground truth image parts, we found that simply using our gradient guidance framework for resampling shows good results. 
  Specifically, in order to guide the initial estimate $\x_T$ to be sufficiently good, we perform $N$ repetition of one reverse sampling $\x_{T-1}$ followed by
one forward step $\x_T =\sqrt{1-\beta_{T-1}}\x_{T-1}+\beta_{T-1}{\bm \epsilon}$ to find the $\x_T$ whose gradient for the next step is easily affected by the loss. With this initial resampling strategy, we can empirically found the initial $\x_T$ that can reduce the number of reverse steps. The overall process is in our algorithm in Appendix.
 % \hspace*{\algorithmicindent}
 % (\hat{x}_0^t,\hat{x}_0^{t+1},x_{src},d_{src},d_{trg})
% \begin{minipage}{\textwidth}
%   \centering

\subsection{Total Loss}
%We use  the losses suggested above to guide the reverse process of the diffusion model. Through the denoised image defined in eq.\ref{eq:4}, 
Putting all together, the final loss in \eqref{eq:total} for the text-guided reverse diffusion  is given by
%the gradient is obtained by calculating the loss function at every intermediate time $t$. Our total loss function for text-guided translation at time $t$ is defined as:
\begin{align}
 \ell_{total} =   \lambda_1 \ell_{cont} + \lambda_2 \ell_{ssim}
 + \lambda_3 \ell_{CLIP} + \lambda_4 \ell_{sem} + \lambda_5 \ell_{rng},
\end{align}
where $\ell_{rng}$ is a regularization loss to prevent the irregular step of reverse diffusion process suggested in (\cite{cgd}). 
If the target style image $\x_{trg}$ is given instead of text conditions $\d_{src}$ and $\d_{trg}$, then
$\ell_{CLIP}$ is simply substituted for $\ell_{sty}$.

\begin{figure}[t!]
\centering
    \includegraphics[width=0.9\linewidth]{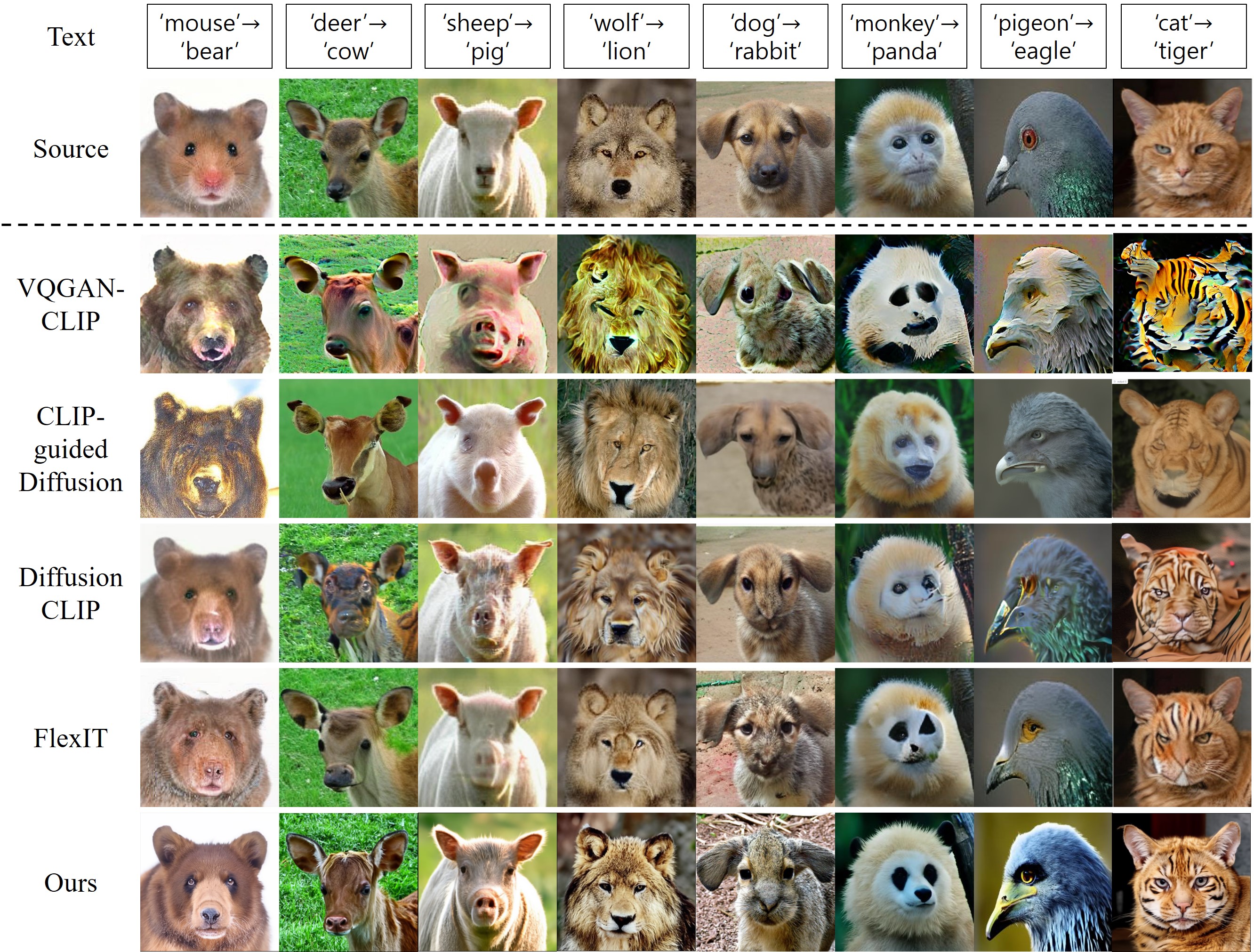}
    \vspace{-0.3cm}
    \caption{Qualitative comparison of text-guided  translation on \textit{Animals} dataset.  Our model generates realistic samples that reflects the  text condition, with better perceptual quality than the baselines.  }
    \label{fig:fig_ani}
    \vspace{-0.3cm}
    \end{figure}

\section{Experiment}
\subsection{Experimental Details}
For implementation, we refer to the official source code of blended diffusion (\cite{blended}). All experiments were performed using unconditional score model pre-trained with Imagenet 256$\times$256 resolution datasets (\cite{diffusionbeat}). In all the experiments, we used diffusion step of $T = 60$ and the resampling repetition of $N=10$; therefore, the total of 70 diffusion reverse steps are used. The generation process takes 40 seconds per image in single RTX 3090 unit. In $\ell_{CLIP}$, we used the ensemble of 5 pre-trained CLIP models (RN50, RN50x4, ViT-B/32, RN50x16, ViT-B/16) for the text-guidance, following the setup of \cite{flexit}. Our detailed experimental settings are elaborated in Appendix.

\subsection{Text-guided Semantic Image Translation}

To evaluate the performance of our text-guided image translation, we conducted comparisons  with state-of-the-art baseline models. For baseline methods, we selected the recently proposed models which use pre-trained CLIP for text-guided image manipulation: VQGAN-CLIP (\cite{vqclip}), CLIP-guided diffusion (CGD) (\cite{cgd}), DiffusionCLIP (\cite{diffusionclip}), and FlexIT (\cite{flexit}). For all baseline methods, we referenced the official source codes.

% For diffusion-based methods(CGD,DiffusionCLIP), we used the same score model (unconditional imagenet 256$\times$256) for fair comparison. In VQGAN-based methods(VQGAN-CLIP,FlexIT), we followed the experimental setting of the original script. 

Since our framework can be applied to arbitrary text semantics, we tried quantitative and qualitative evaluation on various kinds of natural image datasets. We tested our translation performance using two different datasets: animal faces (\cite{animal}) and landscapes (\cite{landscape}). The animal face dataset contains 14 classes of animal face images, and the landscapes dataset consists of 7 classes of various natural landscape images. 

%  For quantitative evaluation on animal face dataset, we calculated the metrics using the outputs of translating the 4 images from a source class into all the remaining classes. Therefore, in our animal face dataset, total of 676 generated images are used for evaluation. For our landscape dataset experiment,
% we translated 5 images from a source class into all the remaining classes,  total of 210 generated images are used for evaluation.

\begin{figure}[t!]
\centering
    \includegraphics[width=0.85\linewidth]{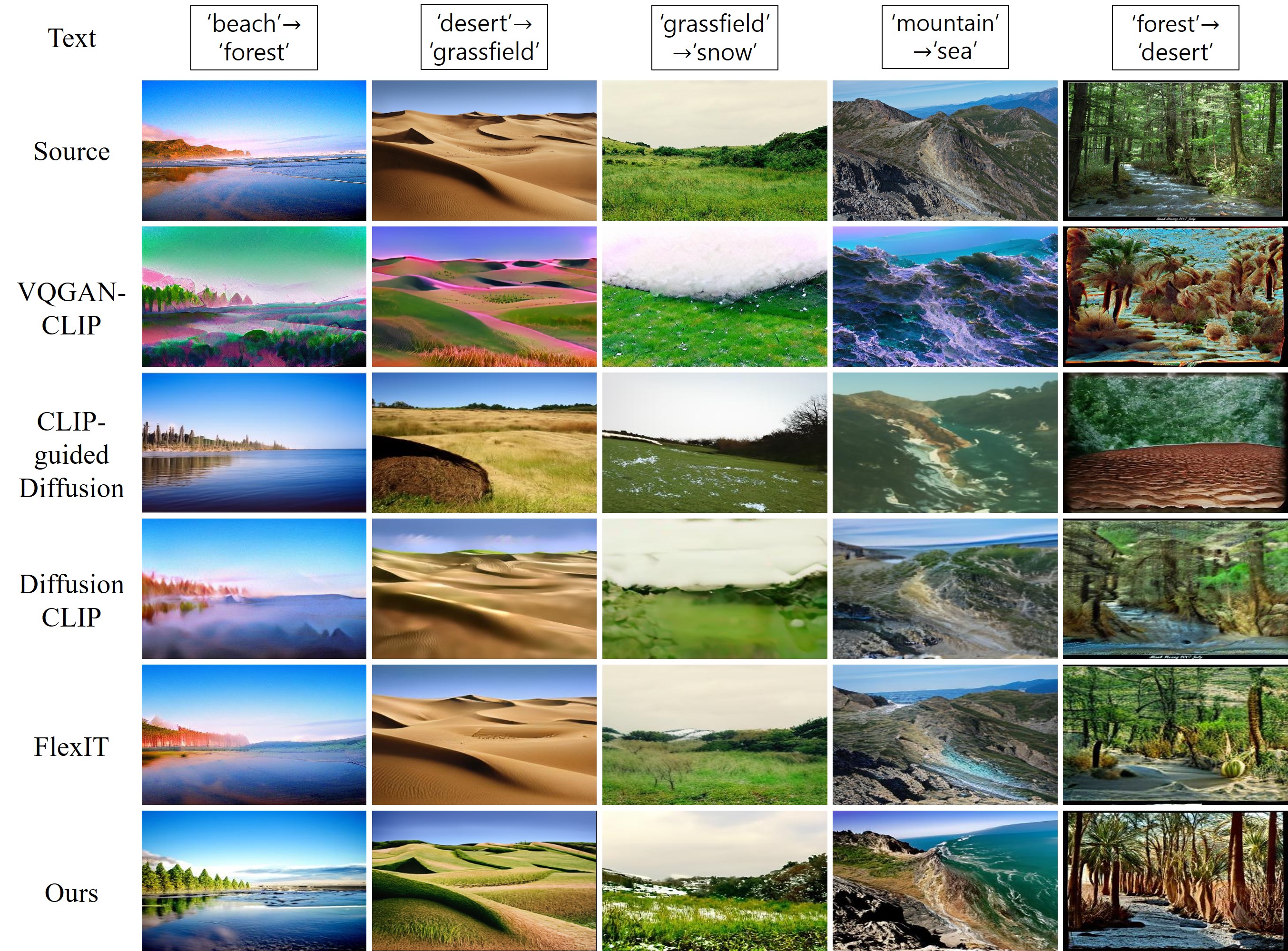}
    \vspace{-0.3cm}
    \caption{Qualitative comparison of text-guided image translation on \textit{Landscape} dataset.  Our model generates outputs with better perceptual quality than the baselines.}
    \label{fig:fig_land}
    \end{figure}
%\vspace{-0.4cm}
\begin{table}[!t]

\begin{center}
\resizebox{1\textwidth}{!}{
\begin{tabular}{@{\extracolsep{5pt}}cccccccccc@{}}
\hline
\multirow{2}{*}{\textbf{Method}}  & \multicolumn{3}{c}{\textbf{Animals}} &\multicolumn{3}{c}{\textbf{Landscapes}}&\multicolumn{3}{c}{\textbf{User Study}}\\

\cline{2-4} 
\cline{5-7}
\cline{8-10}
 & SFID$\downarrow$& CSFID$\downarrow$&LPIPS$\downarrow$& SFID$\downarrow$& CSFID$\downarrow$&LPIPS$\downarrow$  & Text  $\uparrow$ & Realism $\uparrow$& Content $\uparrow$\\

\hline
VQGAN-CLIP &30.01&	65.51 &	 0.462 &	33.31& 82.92  & 0.571 & 2.78&	2.05 &	2.16\\
CLIP-GD &12.50&53.05 &0.468 &18.13& 62.19& 0.458&2.61 & 2.24&2.28 \\
% \cdashline{2-6}
DiffusionCLIP& 25.09& 66.50& 0.379 &29.85&76.29 & 0.568& 2.50&2.54 & 3.06 \\
% \hline
FlexIT &32.71&	57.87 &	0.215 &	18.04 &60.04 & 0.243& 2.22&	3.15 & 3.89 \\
Ours &9.98&	41.07	& 0.372	&16.86& 54.48 & 0.417& 3.68&	4.28	& 4.11	\\

\hline
\end{tabular}
}
% \end{adjustbox}
\end{center}
% }
\vspace*{-.3cm}
\caption{ Quantitative comparison in the text-guided image translation. Our model outperforms baselines in overall scores for both of \textit{Animals} and \textit{Landscapes} datasets as well as user study.}
%\vspace{-0.3cm}
\vspace*{-.3cm}
\label{table:text0}
% \end{threeparttable}
\end{table}

% \begin{table}[t!]
% \begin{minipage}[t]{0.48\textwidth}
% \centering
% \resizebox{1\textwidth}{!}{
% \begin{tabular}{@{\extracolsep{5pt}}cccc@{}}
% \hline
% {\textbf{Method}}  & Text match $\uparrow$ & Realism $\uparrow$& Content $\uparrow$\\

% \hline
% VQGAN-CLIP & 2.78&	2.05 &	2.16  \\
% CLIP-GD&2.61 & 2.24&2.28 \\
% % \cdashline{2-6}
% DiffusionCLIP& 2.50&2.54 & 3.06 \\
% % \hline
% FlexIT & 2.22&	3.15 & 3.89 \\
% Ours & 3.68&	4.28	& 4.11	\\
% \hline
% % \label{table:user_text}
% % \caption{Quantitative comparison of multi-modal image translation.}

% \end{tabular}
% }
% \vspace{-0.3cm}
% \captionof{table}{User study comparison of text-guided image translation tasks. Our model shows better perceptual quality compared to baselines.}
% \label{table:user_text}
% \end{minipage}
% \quad

To measure the performance of the generated images, we measured the FID score (\cite{fid}). However, when using the basic FID score measurement, the output value is not stable because our number of generated images is not large. To compensate for this, we measure the performance using a simplified FID (\cite{sfid}) that does not consider the diagonal term of the feature distributions. Also, we additionally showed a class-wise SFID score that measures the SFID for each class of the converted output because it is necessary to measure whether the converted output accurately reflects the semantic information of the target class. Finally, we used the averaged LPIPS score between input and output to verify the content preservation performance of our method.
% We calculated averaged LPIPS score between output and input source images.  
Further experimental settings can be found in our Appendix.

In Table \ref{table:text0}, we show the quantitative comparison results. In image quality measurement using SFID and CSFID, our model showed the best performance among all baseline methods. Especially for Animals dataset, our SFID value outperformed others in large gain. In the content preservation by  LPIPS score, our method scored the second best. In case of FlexIT, it showed the best score in LPIPS since the model is directly trained with LPIPS loss. However, too low value of LPIPS is undesired as it means that the model failed in proper semantic change. This can be also seen in qualitative result of Figs. \ref{fig:fig_ani} and \ref{fig:fig_land}, where our results have proper semantic features of target texts with content preservation, whereas the results from FlexIT failed in semantic change as it is too strongly confined to the  source images. In other baseline methods, most of the methods failed in proper content preservation. Since our method is based on DDPM, our model can generate diverse images as shown in the additional outputs in our Appendix.

To further evaluate the perceptual quality of generated samples, we conducted user study. In order to measure the detailed opinions, we used custom-made opinion scoring system. We asked the users in three different parts: 1) Are the output have correct semantic of target text? (Text-match), 2) are the generated images realistic? (Realism), 3) do the outputs contains the content information of source images? (Content). Detailed user-study settings are in our Appendix. In Table \ref{table:text0}, our model showed the best performance, which further shows the superiority of our method.

\subsection{Image-guided Semantic Image Translation}
\begin{wraptable}{R}{0.4\textwidth}
% \begin{minipage}[t]{0.49\textwidth}
\centering
\resizebox{0.4\textwidth}{!}{
\begin{tabular}{@{\extracolsep{5pt}}cccc@{}}
\hline
{\textbf{Method}}  & Style  $\uparrow$ & Realism $\uparrow$& Content $\uparrow$\\

\hline

SANet & 2.75 &	4.08 &	4.37  \\
WCT2 & 2.59& 4.64& 4.90\\
% \cdashline{2-6}
STROTSS& 3.92& 2.91 & 3.17 \\
% \hline
SplicingViT & 3.50&	2.08 &  2.15\\
Ours & 4.23&	4.25	& 	4.51\\
% \cdashline{2-6}
 % & \textbf{54.02}&	\textbf{0.470}&	\textbf{56.91}&	\textbf{0.362}\\
\hline

\end{tabular}
}
\vspace{-0.3cm}
\captionof{table}{User study comparison of image-guided  translation tasks. Our model outperforms baseline methods in overall perceptual quality.}
\label{table:user_image}
\vspace{-0.3cm}
% \end{minipage}
\end{wraptable}

Since our method can be easily adapted to the image translation guided by target images, we evaluate the performance of our model with comparison experiments. We compare our model with appearance transfer models of Splicing ViT (\cite{splice}), STROTSS (\cite{strotss}), and style transfer methods WCT2 (\cite{wct2}) and SANet (\cite{sanet}). 

\begin{figure}[t!]
\centering
    \includegraphics[width=0.9\linewidth]{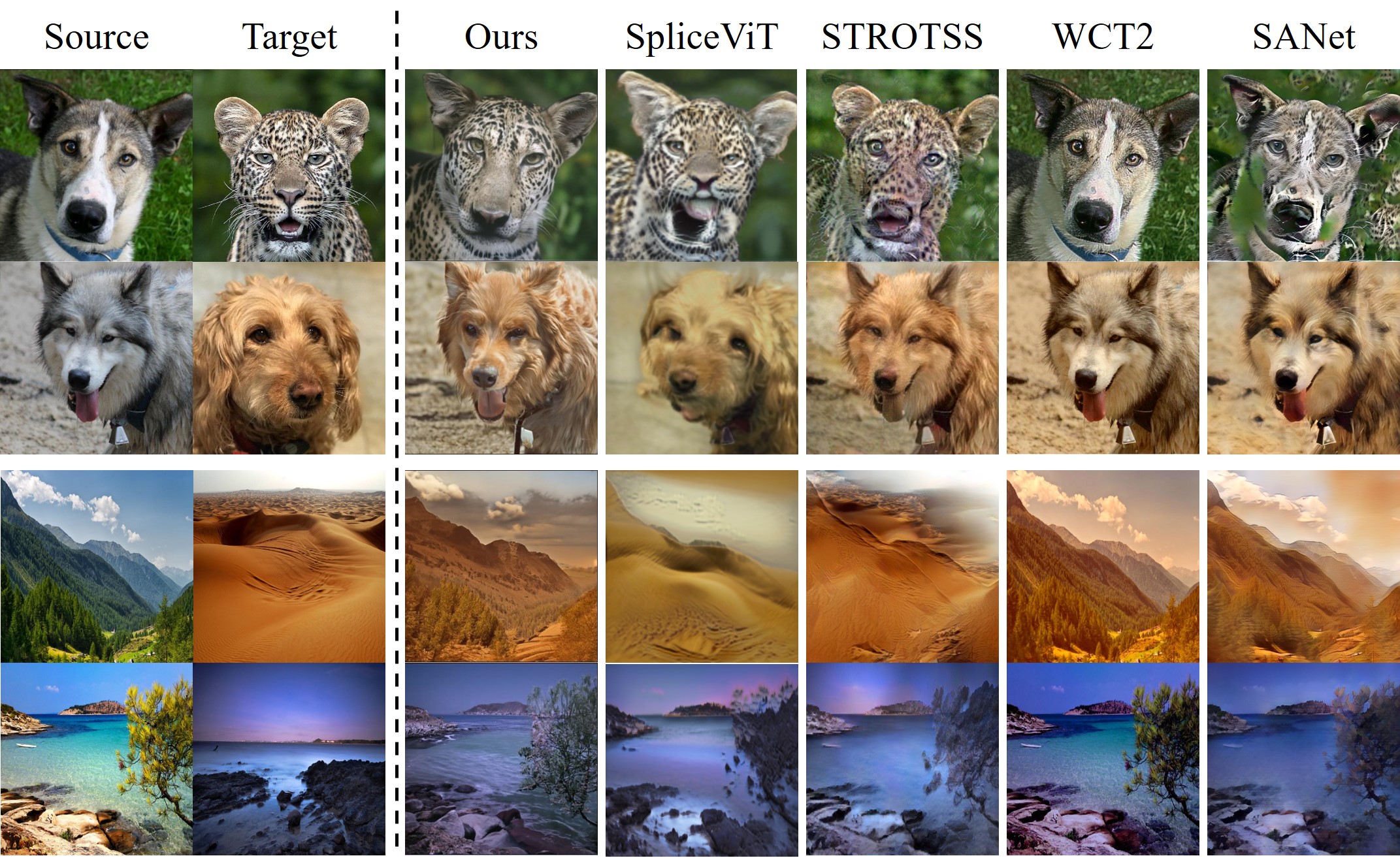}
    \vspace{-0.3cm}
    \caption{Qualitative comparison of image-guided image translation.  Our results have better perceptual quality than the baseline outputs.}
    \label{fig:fig_img}
    \vspace{-0.4cm}
    \end{figure}

Fig.~\ref{fig:fig_img} is a qualitative comparison result of image guided translation task. Our model successfully generated outputs that follow the semantic styles of the target images while maintaining the content of the source images. In the case of other models, we can see that the content was severely deformed or the semantic style was not properly reflected. We also measured the overall perceptual quality through a user study. As with text-guided translation, we investigated user opinion through three different questions. In Table \ref{table:user_image}, our model obtained the best score in style matching score and the second best in realism and content preservation scores. Baseline WCT2 showed the best in realism and content scores, but it shows the worst score in style matching because the outputs are hardly changed from the inputs except for overall colors. The opinions scores confirm that our model outperforms the baselines. More details are in our Appendix.

\subsection{Ablation Study}
\begin{figure}[t!]
\centering
    \includegraphics[width=0.9\linewidth]{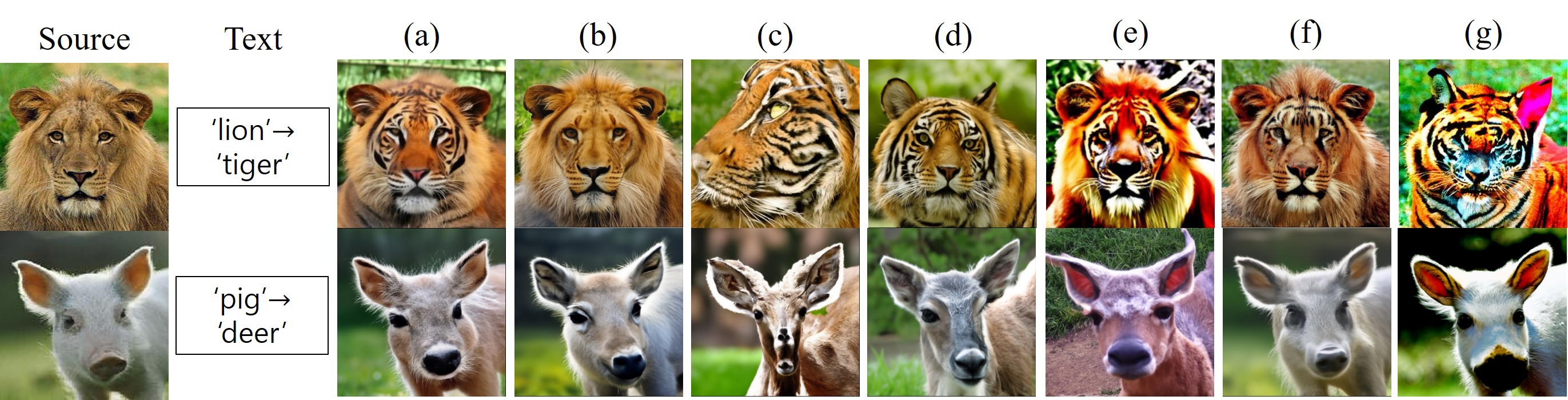}
    \vspace{-0.3cm}
    \caption{Qualitative comparison on ablation study. Our full setting shows the best results.
    % (a) The results from our best setting, (b) outputs without $L_{sem}$, (c) outputs without $L_{con}$, (d) samples using LPIPS loss instead of $L_{con}$, (e) samples using L2 maximization instead of $L_{sem}$, (f) outputs withour proposed resampling strategy. Our best settings shows the best perceptual quality.  
    }
    \label{fig:fig_abl}
    \vspace{-0.4cm}
    \end{figure}

To verify the proposed components in our framework, we compare the generation performance with different settings. In Fig.~\ref{fig:fig_abl}, we show that (a) the outputs from our best setting have the correct semantic of target text, with preserving the content of the source; (b) by removing $\ell_{sem}$, the results still have the appearance of source images, suggesting that images are not fully converted to the target domain; (c) without $\ell_{cont}$, the output images totally failed to capture the content of source images; (d)  by using LPIPS perceptual loss instead of  proposed $\ell_{cont}$, the results can only capture the approximate content of source images; (e) using pixel-wise $l_2$ maximization loss instead of proposed $\ell_{sem}$, the outputs suffer from irregular artifacts; (f) without using our proposed resampling trick, the results cannot fully reflect the semantic information of target texts. {(g) With using VGG16 network instead of DINO ViT, the output structure is severely degraded with artifacts.} Overall, we can obtain the best generation outputs by using all of our proposed components. For further evaluation, we will show the quantitative results of ablation study in our Appendix.

\section{Conclusion}

In conclusion, we proposed a novel loss function which utilizes a pre-trained ViT model to guide the generation process of DDPM models in terms of content preservation and semantic changes. We further propose a novel strategy of resampling technique for better initialization of diffusion process. For evaluation, our extensive experimental results show that our proposed framework has superior performance compared to baselines in both of text- and image-guided semantic image translation tasks. Despite the successful results, our method often fails to translate the image styles when there is large domain gap between source and target. With respect to this, we show the failure cases and discussions on limitations in Appendix. 

{\large\noindent\textbf{Acknowledgement}}

This research was supported by Field-oriented Technology Development Project for Customs Administration through National Research Foundation of Korea(NRF) funded by the Ministry of Science \& ICT and Korea Customs Service(NRF-2021M3I1A1097938), the KAIST Key Research Institute (Interdisciplinary Research Group) Project, and the National Research Foundation of Korea under Grant
(NRF-2020R1A2B5B03001980).

\bibliography{iclr2023_conference}

\begin{thebibliography}{57}
\providecommand{\natexlab}[1]{#1}
\providecommand{\url}[1]{\texttt{#1}}
\expandafter\ifx\csname urlstyle\endcsname\relax
  \providecommand{\doi}[1]{doi: #1}\else
  \providecommand{\doi}{doi: \begingroup \urlstyle{rm}\Url}\fi

\bibitem[Anoosheh et~al.(2018)Anoosheh, Agustsson, Timofte, and
  Van~Gool]{seasons}
Asha Anoosheh, Eirikur Agustsson, Radu Timofte, and Luc Van~Gool.
\newblock Combogan: Unrestrained scalability for image domain translation.
\newblock In \emph{Proceedings of the IEEE Conference on Computer Vision and
  Pattern Recognition Workshops}, pp.\  783--790, 2018.

\bibitem[Avrahami et~al.(2022)Avrahami, Lischinski, and Fried]{blended}
Omri Avrahami, Dani Lischinski, and Ohad Fried.
\newblock Blended diffusion for text-driven editing of natural images.
\newblock In \emph{Proceedings of the IEEE/CVF Conference on Computer Vision
  and Pattern Recognition}, pp.\  18208--18218, 2022.

\bibitem[Caron et~al.(2021)Caron, Touvron, Misra, J{\'e}gou, Mairal,
  Bojanowski, and Joulin]{dino}
Mathilde Caron, Hugo Touvron, Ishan Misra, Herv{\'e} J{\'e}gou, Julien Mairal,
  Piotr Bojanowski, and Armand Joulin.
\newblock Emerging properties in self-supervised vision transformers.
\newblock In \emph{Proceedings of the IEEE/CVF International Conference on
  Computer Vision}, pp.\  9650--9660, 2021.

\bibitem[Chen et~al.(2018)Chen, Lai, and Liu]{landscape}
Yang Chen, Yu-Kun Lai, and Yong-Jin Liu.
\newblock Cartoongan: Generative adversarial networks for photo cartoonization.
\newblock In \emph{Proceedings of the IEEE conference on computer vision and
  pattern recognition}, pp.\  9465--9474, 2018.

\bibitem[Choi et~al.(2018)Choi, Choi, Kim, Ha, Kim, and Choo]{stargan}
Yunjey Choi, Minje Choi, Munyoung Kim, Jung-Woo Ha, Sunghun Kim, and Jaegul
  Choo.
\newblock Stargan: Unified generative adversarial networks for multi-domain
  image-to-image translation.
\newblock In \emph{Proceedings of the IEEE conference on computer vision and
  pattern recognition}, pp.\  8789--8797, 2018.

\bibitem[Chong \& Forsyth(2021)Chong and Forsyth]{jojogan}
Min~Jin Chong and David Forsyth.
\newblock Jojogan: One shot face stylization.
\newblock \emph{arXiv preprint arXiv:2112.11641}, 2021.

\bibitem[Chung et~al.(2022{\natexlab{a}})Chung, Sim, Ryu, and Ye]{mcg}
Hyungjin Chung, Byeongsu Sim, Dohoon Ryu, and Jong~Chul Ye.
\newblock Improving diffusion models for inverse problems using manifold
  constraints.
\newblock \emph{arXiv preprint arXiv:2206.00941}, 2022{\natexlab{a}}.

\bibitem[Chung et~al.(2022{\natexlab{b}})Chung, Sim, and Ye]{ccdf}
Hyungjin Chung, Byeongsu Sim, and Jong~Chul Ye.
\newblock Come-closer-diffuse-faster: Accelerating conditional diffusion models
  for inverse problems through stochastic contraction.
\newblock In \emph{Proceedings of the IEEE/CVF Conference on Computer Vision
  and Pattern Recognition}, pp.\  12413--12422, 2022{\natexlab{b}}.

\bibitem[Couairon et~al.(2022)Couairon, Grechka, Verbeek, Schwenk, and
  Cord]{flexit}
Guillaume Couairon, Asya Grechka, Jakob Verbeek, Holger Schwenk, and Matthieu
  Cord.
\newblock Flexit: Towards flexible semantic image translation.
\newblock In \emph{Proceedings of the IEEE/CVF Conference on Computer Vision
  and Pattern Recognition}, pp.\  18270--18279, 2022.

\bibitem[Crowson(2022)]{cgd}
Katherine Crowson.
\newblock Clip-guided diffusion.
\newblock 2022.
\newblock URL \url{https://github.com/afiaka87/clip-guided-diffusion}.

\bibitem[Crowson et~al.(2022)Crowson, Biderman, Kornis, Stander, Hallahan,
  Castricato, and Raff]{vqclip}
Katherine Crowson, Stella Biderman, Daniel Kornis, Dashiell Stander, Eric
  Hallahan, Louis Castricato, and Edward Raff.
\newblock Vqgan-clip: Open domain image generation and editing with natural
  language guidance.
\newblock \emph{arXiv preprint arXiv:2204.08583}, 2022.

\bibitem[Deng et~al.(2019)Deng, Guo, Xue, and Zafeiriou]{arcface}
Jiankang Deng, Jia Guo, Niannan Xue, and Stefanos Zafeiriou.
\newblock Arcface: Additive angular margin loss for deep face recognition.
\newblock In \emph{Proceedings of the IEEE/CVF conference on computer vision
  and pattern recognition}, pp.\  4690--4699, 2019.

\bibitem[Dhariwal \& Nichol(2021)Dhariwal and Nichol]{diffusionbeat}
Prafulla Dhariwal and Alexander Nichol.
\newblock Diffusion models beat gans on image synthesis.
\newblock \emph{Advances in Neural Information Processing Systems},
  34:\penalty0 8780--8794, 2021.

\bibitem[Dosovitskiy et~al.(2020)Dosovitskiy, Beyer, Kolesnikov, Weissenborn,
  Zhai, Unterthiner, Dehghani, Minderer, Heigold, Gelly, et~al.]{vit}
Alexey Dosovitskiy, Lucas Beyer, Alexander Kolesnikov, Dirk Weissenborn,
  Xiaohua Zhai, Thomas Unterthiner, Mostafa Dehghani, Matthias Minderer, Georg
  Heigold, Sylvain Gelly, et~al.
\newblock An image is worth 16x16 words: Transformers for image recognition at
  scale.
\newblock \emph{arXiv preprint arXiv:2010.11929}, 2020.

\bibitem[Esser et~al.(2021)Esser, Rombach, and Ommer]{vqgan}
Patrick Esser, Robin Rombach, and Bjorn Ommer.
\newblock Taming transformers for high-resolution image synthesis.
\newblock In \emph{Proceedings of the IEEE/CVF conference on computer vision
  and pattern recognition}, pp.\  12873--12883, 2021.

\bibitem[Fu et~al.(2021)Fu, Wang, and Wang]{ldast}
Tsu-Jui Fu, Xin~Eric Wang, and William~Yang Wang.
\newblock Language-driven image style transfer.
\newblock 2021.

\bibitem[Gal et~al.(2021)Gal, Patashnik, Maron, Chechik, and Cohen-Or]{nada}
Rinon Gal, Or~Patashnik, Haggai Maron, Gal Chechik, and Daniel Cohen-Or.
\newblock Stylegan-nada: Clip-guided domain adaptation of image generators.
\newblock \emph{arXiv preprint arXiv:2108.00946}, 2021.

\bibitem[Gatys et~al.(2016)Gatys, Ecker, and Bethge]{nst}
Leon~A Gatys, Alexander~S Ecker, and Matthias Bethge.
\newblock Image style transfer using convolutional neural networks.
\newblock In \emph{Proceedings of the IEEE conference on computer vision and
  pattern recognition}, pp.\  2414--2423, 2016.

\bibitem[Granot et~al.(2022)Granot, Feinstein, Shocher, Bagon, and Irani]{gpen}
Niv Granot, Ben Feinstein, Assaf Shocher, Shai Bagon, and Michal Irani.
\newblock Drop the gan: In defense of patches nearest neighbors as single image
  generative models.
\newblock In \emph{Proceedings of the IEEE/CVF Conference on Computer Vision
  and Pattern Recognition}, pp.\  13460--13469, 2022.

\bibitem[Hahne \& Aggoun(2021)Hahne and Aggoun]{color}
Christopher Hahne and Amar Aggoun.
\newblock Plenopticam v1.0: A light-field imaging framework.
\newblock \emph{IEEE Transactions on Image Processing}, 30:\penalty0
  6757--6771, 2021.
\newblock \doi{10.1109/TIP.2021.3095671}.

\bibitem[Heusel et~al.(2017)Heusel, Ramsauer, Unterthiner, Nessler, and
  Hochreiter]{fid}
Martin Heusel, Hubert Ramsauer, Thomas Unterthiner, Bernhard Nessler, and Sepp
  Hochreiter.
\newblock Gans trained by a two time-scale update rule converge to a local nash
  equilibrium.
\newblock pp.\  6626--6637, 2017.

\bibitem[Ho et~al.(2020)Ho, Jain, and Abbeel]{ddpm}
Jonathan Ho, Ajay Jain, and Pieter Abbeel.
\newblock Denoising diffusion probabilistic models.
\newblock \emph{Advances in Neural Information Processing Systems},
  33:\penalty0 6840--6851, 2020.

\bibitem[Huang \& Belongie(2017)Huang and Belongie]{adain}
Xun Huang and Serge Belongie.
\newblock Arbitrary style transfer in real-time with adaptive instance
  normalization.
\newblock pp.\  1501--1510, 2017.

\bibitem[Isola et~al.(2017)Isola, Zhu, Zhou, and Efros]{pix2pix}
Phillip Isola, Jun-Yan Zhu, Tinghui Zhou, and Alexei~A Efros.
\newblock Image-to-image translation with conditional adversarial networks.
\newblock In \emph{Proceedings of the IEEE conference on computer vision and
  pattern recognition}, pp.\  1125--1134, 2017.

\bibitem[Jung et~al.(2022)Jung, Kwon, and Ye]{HnegSRC}
Chanyong Jung, Gihyun Kwon, and Jong~Chul Ye.
\newblock Exploring patch-wise semantic relation for contrastive learning in
  image-to-image translation tasks.
\newblock In \emph{Proceedings of the IEEE/CVF Conference on Computer Vision
  and Pattern Recognition (CVPR)}, pp.\  18260--18269, June 2022.

\bibitem[Karras et~al.(2020)Karras, Laine, Aittala, Hellsten, Lehtinen, and
  Aila]{stylegan2}
Tero Karras, Samuli Laine, Miika Aittala, Janne Hellsten, Jaakko Lehtinen, and
  Timo Aila.
\newblock Analyzing and improving the image quality of stylegan.
\newblock In \emph{Proceedings of the IEEE/CVF conference on computer vision
  and pattern recognition}, pp.\  8110--8119, 2020.

\bibitem[Kim et~al.(2020)Kim, Kim, Jung, and Hwang]{sfid}
Chung-Il Kim, Meejoung Kim, Seungwon Jung, and Eenjun Hwang.
\newblock Simplified fréchet distance for generative adversarial nets.
\newblock \emph{Sensors}, 20\penalty0 (6), 2020.
\newblock ISSN 1424-8220.
\newblock \doi{10.3390/s20061548}.
\newblock URL \url{https://www.mdpi.com/1424-8220/20/6/1548}.

\bibitem[Kim et~al.(2022)Kim, Kwon, and Ye]{diffusionclip}
Gwanghyun Kim, Taesung Kwon, and Jong~Chul Ye.
\newblock Diffusionclip: Text-guided diffusion models for robust image
  manipulation.
\newblock In \emph{Proceedings of the IEEE/CVF Conference on Computer Vision
  and Pattern Recognition}, pp.\  2426--2435, 2022.

\bibitem[Kim \& Ye(2021)Kim and Ye]{noise2score}
Kwanyoung Kim and Jong~Chul Ye.
\newblock Noise2score: tweedie’s approach to self-supervised image denoising
  without clean images.
\newblock \emph{Advances in Neural Information Processing Systems},
  34:\penalty0 864--874, 2021.

\bibitem[Kolkin et~al.(2019)Kolkin, Salavon, and Shakhnarovich]{strotss}
Nicholas Kolkin, Jason Salavon, and Gregory Shakhnarovich.
\newblock Style transfer by relaxed optimal transport and self-similarity.
\newblock In \emph{Proceedings of the IEEE/CVF Conference on Computer Vision
  and Pattern Recognition}, pp.\  10051--10060, 2019.

\bibitem[Kwon \& Ye(2021)Kwon and Ye]{clipstyler}
Gihyun Kwon and Jong~Chul Ye.
\newblock Clipstyler: Image style transfer with a single text condition.
\newblock \emph{arXiv preprint arXiv:2112.00374}, 2021.

\bibitem[Kwon \& Ye(2022)Kwon and Ye]{oneshotclip}
Gihyun Kwon and Jong~Chul Ye.
\newblock One-shot adaptation of gan in just one clip.
\newblock \emph{arXiv preprint arXiv:2203.09301}, 2022.

\bibitem[Lee et~al.(2019)Lee, Kim, Moon, and Ye]{collagan}
Dongwook Lee, Junyoung Kim, Won-Jin Moon, and Jong~Chul Ye.
\newblock Collagan: Collaborative gan for missing image data imputation.
\newblock In \emph{Proceedings of the IEEE/CVF Conference on Computer Vision
  and Pattern Recognition}, pp.\  2487--2496, 2019.

\bibitem[Lin et~al.(2020)Lin, Pang, Xia, Chen, and Luo]{tuigan}
Jianxin Lin, Yingxue Pang, Yingce Xia, Zhibo Chen, and Jiebo Luo.
\newblock Tuigan: Learning versatile image-to-image translation with two
  unpaired images.
\newblock In \emph{European Conference on Computer Vision}, pp.\  18--35.
  Springer, 2020.

\bibitem[Liu et~al.(2021)Liu, Gong, Wu, Zhang, Su, and Liu]{fusedream}
Xingchao Liu, Chengyue Gong, Lemeng Wu, Shujian Zhang, Hao Su, and Qiang Liu.
\newblock Fusedream: Training-free text-to-image generation with improved clip+
  gan space optimization.
\newblock \emph{arXiv preprint arXiv:2112.01573}, 2021.

\bibitem[L{\"u}ddecke \& Ecker(2022)L{\"u}ddecke and Ecker]{clipseg}
Timo L{\"u}ddecke and Alexander Ecker.
\newblock Image segmentation using text and image prompts.
\newblock In \emph{Proceedings of the IEEE/CVF Conference on Computer Vision
  and Pattern Recognition}, pp.\  7086--7096, 2022.

\bibitem[Nichol \& Dhariwal(2021)Nichol and Dhariwal]{iddpm}
Alexander~Quinn Nichol and Prafulla Dhariwal.
\newblock Improved denoising diffusion probabilistic models.
\newblock In \emph{International Conference on Machine Learning}, pp.\
  8162--8171. PMLR, 2021.

\bibitem[Ojha et~al.(2021)Ojha, Li, Lu, Efros, Lee, Shechtman, and Zhang]{fsga}
Utkarsh Ojha, Yijun Li, Jingwan Lu, Alexei~A Efros, Yong~Jae Lee, Eli
  Shechtman, and Richard Zhang.
\newblock Few-shot image generation via cross-domain correspondence.
\newblock In \emph{Proceedings of the IEEE/CVF Conference on Computer Vision
  and Pattern Recognition}, pp.\  10743--10752, 2021.

\bibitem[Park \& Lee(2019)Park and Lee]{sanet}
Dae~Young Park and Kwang~Hee Lee.
\newblock Arbitrary style transfer with style-attentional networks.
\newblock In \emph{Proceedings of the IEEE/CVF Conference on Computer Vision
  and Pattern Recognition}, pp.\  5880--5888, 2019.

\bibitem[Park et~al.(2020)Park, Efros, Zhang, and Zhu]{cut}
Taesung Park, Alexei~A. Efros, Richard Zhang, and Jun-Yan Zhu.
\newblock Contrastive learning for unpaired image-to-image translation.
\newblock In Andrea Vedaldi, Horst Bischof, Thomas Brox, and Jan-Michael Frahm
  (eds.), \emph{Computer Vision -- ECCV 2020}, pp.\  319--345, Cham, 2020.
  Springer International Publishing.
\newblock ISBN 978-3-030-58545-7.

\bibitem[Patashnik et~al.(2021)Patashnik, Wu, Shechtman, Cohen-Or, and
  Lischinski]{styleclip}
Or~Patashnik, Zongze Wu, Eli Shechtman, Daniel Cohen-Or, and Dani Lischinski.
\newblock Styleclip: Text-driven manipulation of stylegan imagery.
\newblock In \emph{Proceedings of the IEEE/CVF International Conference on
  Computer Vision}, pp.\  2085--2094, 2021.

\bibitem[Radford et~al.(2021)Radford, Kim, Hallacy, Ramesh, Goh, Agarwal,
  Sastry, Askell, Mishkin, Clark, et~al.]{clip_rad}
Alec Radford, Jong~Wook Kim, Chris Hallacy, Aditya Ramesh, Gabriel Goh,
  Sandhini Agarwal, Girish Sastry, Amanda Askell, Pamela Mishkin, Jack Clark,
  et~al.
\newblock Learning transferable visual models from natural language
  supervision.
\newblock \emph{arXiv preprint arXiv:2103.00020}, 2021.

\bibitem[Ramesh et~al.(2022)Ramesh, Dhariwal, Nichol, Chu, and Chen]{dalle2}
Aditya Ramesh, Prafulla Dhariwal, Alex Nichol, Casey Chu, and Mark Chen.
\newblock Hierarchical text-conditional image generation with clip latents.
\newblock \emph{arXiv preprint arXiv:2204.06125}, 2022.

\bibitem[Saharia et~al.(2022{\natexlab{a}})Saharia, Chan, Chang, Lee, Ho,
  Salimans, Fleet, and Norouzi]{saharia2022palette}
Chitwan Saharia, William Chan, Huiwen Chang, Chris Lee, Jonathan Ho, Tim
  Salimans, David Fleet, and Mohammad Norouzi.
\newblock Palette: Image-to-image diffusion models.
\newblock In \emph{ACM SIGGRAPH 2022 Conference Proceedings}, pp.\  1--10,
  2022{\natexlab{a}}.

\bibitem[Saharia et~al.(2022{\natexlab{b}})Saharia, Chan, Saxena, Li, Whang,
  Denton, Ghasemipour, Ayan, Mahdavi, Lopes, et~al.]{imagen}
Chitwan Saharia, William Chan, Saurabh Saxena, Lala Li, Jay Whang, Emily
  Denton, Seyed Kamyar~Seyed Ghasemipour, Burcu~Karagol Ayan, S~Sara Mahdavi,
  Rapha~Gontijo Lopes, et~al.
\newblock Photorealistic text-to-image diffusion models with deep language
  understanding.
\newblock \emph{arXiv preprint arXiv:2205.11487}, 2022{\natexlab{b}}.

\bibitem[Sasaki et~al.(2021)Sasaki, Willcocks, and Breckon]{unitddpm}
Hiroshi Sasaki, Chris~G Willcocks, and Toby~P Breckon.
\newblock Unit-ddpm: Unpaired image translation with denoising diffusion
  probabilistic models.
\newblock \emph{arXiv preprint arXiv:2104.05358}, 2021.

\bibitem[Si \& Zhu(2012)Si and Zhu]{animal}
Zhangzhang Si and Song-Chun Zhu.
\newblock Learning hybrid image templates (hit) by information projection.
\newblock \emph{IEEE Transactions on Pattern Analysis and Machine
  Intelligence}, 34\penalty0 (7):\penalty0 1354--1367, 2012.
\newblock \doi{10.1109/TPAMI.2011.227}.

\bibitem[Song et~al.(2020{\natexlab{a}})Song, Meng, and
  Ermon]{song2020denoising}
Jiaming Song, Chenlin Meng, and Stefano Ermon.
\newblock Denoising diffusion implicit models.
\newblock In \emph{International Conference on Learning Representations},
  2020{\natexlab{a}}.

\bibitem[Song et~al.(2020{\natexlab{b}})Song, Sohl-Dickstein, Kingma, Kumar,
  Ermon, and Poole]{scoresde}
Yang Song, Jascha Sohl-Dickstein, Diederik~P Kingma, Abhishek Kumar, Stefano
  Ermon, and Ben Poole.
\newblock Score-based generative modeling through stochastic differential
  equations.
\newblock \emph{arXiv preprint arXiv:2011.13456}, 2020{\natexlab{b}}.

\bibitem[Tumanyan et~al.(2022)Tumanyan, Bar-Tal, Bagon, and Dekel]{splice}
Narek Tumanyan, Omer Bar-Tal, Shai Bagon, and Tali Dekel.
\newblock Splicing vit features for semantic appearance transfer.
\newblock In \emph{Proceedings of the IEEE/CVF Conference on Computer Vision
  and Pattern Recognition}, pp.\  10748--10757, 2022.

\bibitem[Wang et~al.(2022{\natexlab{a}})Wang, Chai, He, Chen, and
  Liao]{clipnerf}
Can Wang, Menglei Chai, Mingming He, Dongdong Chen, and Jing Liao.
\newblock Clip-nerf: Text-and-image driven manipulation of neural radiance
  fields.
\newblock In \emph{Proceedings of the IEEE/CVF Conference on Computer Vision
  and Pattern Recognition}, pp.\  3835--3844, 2022{\natexlab{a}}.

\bibitem[Wang et~al.(2022{\natexlab{b}})Wang, Lu, Li, Tao, Guo, Gong, and
  Liu]{cris}
Zhaoqing Wang, Yu~Lu, Qiang Li, Xunqiang Tao, Yandong Guo, Mingming Gong, and
  Tongliang Liu.
\newblock Cris: Clip-driven referring image segmentation.
\newblock In \emph{Proceedings of the IEEE/CVF Conference on Computer Vision
  and Pattern Recognition}, pp.\  11686--11695, 2022{\natexlab{b}}.

\bibitem[Wei et~al.(2022)Wei, Chen, Zhou, Liao, Tan, Yuan, Zhang, and
  Yu]{hairclip}
Tianyi Wei, Dongdong Chen, Wenbo Zhou, Jing Liao, Zhentao Tan, Lu~Yuan, Weiming
  Zhang, and Nenghai Yu.
\newblock Hairclip: Design your hair by text and reference image.
\newblock In \emph{Proceedings of the IEEE/CVF Conference on Computer Vision
  and Pattern Recognition}, pp.\  18072--18081, 2022.

\bibitem[Yoo et~al.(2019)Yoo, Uh, Chun, Kang, and Ha]{wct2}
Jaejun Yoo, Youngjung Uh, Sanghyuk Chun, Byeongkyu Kang, and Jung-Woo Ha.
\newblock Photorealistic style transfer via wavelet transforms.
\newblock In \emph{Proceedings of the IEEE/CVF International Conference on
  Computer Vision}, pp.\  9036--9045, 2019.

\bibitem[Zhou et~al.(2017)Zhou, Zhao, Puig, Fidler, Barriuso, and
  Torralba]{segmentation}
Bolei Zhou, Hang Zhao, Xavier Puig, Sanja Fidler, Adela Barriuso, and Antonio
  Torralba.
\newblock Scene parsing through ade20k dataset.
\newblock In \emph{Proceedings of the IEEE Conference on Computer Vision and
  Pattern Recognition}, 2017.

\bibitem[Zhu et~al.(2017)Zhu, Park, Isola, and Efros]{cyclegan}
Jun-Yan Zhu, Taesung Park, Phillip Isola, and Alexei~A. Efros.
\newblock Unpaired image-to-image translation using cycle-consistent
  adversarial networks.
\newblock In \emph{Proceedings of the IEEE International Conference on Computer
  Vision (ICCV)}, Oct 2017.

\bibitem[Zhu et~al.(2021)Zhu, Abdal, Femiani, and Wonka]{mind}
Peihao Zhu, Rameen Abdal, John Femiani, and Peter Wonka.
\newblock Mind the gap: Domain gap control for single shot domain adaptation
  for generative adversarial networks.
\newblock \emph{arXiv preprint arXiv:2110.08398}, 2021.

\end{thebibliography}
\bibliographystyle{iclr2023_conference}

\newpage
\appendix
\section{Experimental Details}
\subsection{Implementation Details}
For implementation, in case of using text-guided image manipulation, our initial sampling numbers are set as $T=100$, but we skipped the initial 40 steps to maintain the abstract content of input image. Therefore, the total number of sampling steps is $T=60$. With resampling step of $N=10$, we use total of 70 iterations for single image output. We found that using more resampling steps does not show meaningful performance improvement. In image-guided manipulation, we set initial sampling number $T=200$, and skipped the initial 80 steps. We used resampling step $N=10$. Therefore, we use total of 130 iterations. Although we used more iterations than text-guided translation, it takes about 40 seconds.

For hyperparameters, we use $\lambda_1=200$, $\lambda_2=100$, $\lambda_3=2000$,  $\lambda_4=1000$, $\lambda_5=200$. For image-guided translation, we set $\lambda_{mse}=1.5$. For our CLIP loss, we set $\lambda_s=0.4$ and $\lambda_i=0.2$. For our ViT backbone model, we used pre-trained DINO ViT that follows the baseline of Splicing ViT \citep{splice}. For extracting keys of intermediate layer, we use layer of $l=11$, and for [CLS] token, we used last layer output. Since ViT and CLIP model only take 224$\times$224 resolution images, we resized all images before calculating the losses with ViT and CLIP. 

To further improve the sample quality of our qualitative results, we used restarting trick in which we check the $\ell_{reg}$ loss calculated at initial time step $T$, and restart the whole process if the loss value is too high. If the initial loss $\ell_{reg}>0.01$, we restarted the process. For quantitative result, we did not use the restart trick for fair comparison. 

For augmentation, we use the same geometrical augmentations proposed in FlexIT\citep{flexit}. Also, following the setting from CLIP-guided diffusion\citep{cgd}, we included noise augmentation in which we mix the noisy image to $\hat{x}_0(x_t)$ as it further removes the artifacts.

In our image-guided image translation on natural landscape images, we matched the color distribution of output image to that of target image with \citep{color}, as it showed better perceptual quality. Our detailed implementation can be found in our official GitHub repository.\footnote{\url{https://github.com/anon294384/DiffuseIT}}

For baseline experiments, we followed the official source codes in all of the models\footnote{\url{https://github.com/afiaka87/clip-guided-diffusion}}\footnote{\url{https://github.com/nerdyrodent/VQGAN-CLIP}}\footnote{\url{https://github.com/gwang-kim/DiffusionCLIP}}\footnote{\url{https://github.com/facebookresearch/SemanticImageTranslation}}. For diffusion-based models (DiffusionCLIP, CLIP-guided diffusion), we used unconditional score model pre-trained on 256$\times$256 resolutions. In DiffusionCLIP, we fine-tuned the score model longer than suggested training iteration, as it showed better quality. In CLIP-guided diffusion, we set the CLIP-guided loss as 2000, and also set initial sampling number as $T=100$ with skipping initial 40 steps. For VQGAN-based models (FlexIT, VQGAN-CLIP), we used VQGAN trained on imagenet 256$\times$256 resolutions datasets. In VQGAN-CLIP, as using longer iteration results in extremely degraded images, therefore we optimized only 30 iterations, which is smaller than suggested iterations ( $\ge$80). In the experiments of FlexIT, we followed the exactly same settings suggested in the original paper. 

For baselines of image-guided style transfer tasks, we also referenced the original source codes\footnote{\url{https://github.com/omerbt/Splice}}\footnote{\url{https://github.com/nkolkin13/STROTSS}}\footnote{\url{https://github.com/clovaai/WCT2}}\footnote{\url{https://github.com/GlebSBrykin/SANET}}. In all of the experiments, we followed the suggested settings from the original papers.

\subsection{Dataset Details}
For our quantitative results using text-guided image translation, we used two different datasets \textit{Animals} and \textit{Landscapes}. In Animals dataset, the original dataset contains 21 different classes, but we filtered out the images from 14 classes (bear, cat, cow, deer, dog, lion, monkey, mouse, panda, pig, rabbit, sheep, tiger, wolf) which can be classified as mammals. Remaining classes (e.g. human, chicken, etc.) are removed since they have far different semantics from the mammal faces.%; therefore the output image quality in severely unstable in our baseline models. 
Therefore we reported  quantitative scores only with filtered datasets for fair comparison. The dataset contains 100-300 images per each class, and we selected 4 testing images from each class in order to use them as our content source images.  With selected samples, we calculated the metrics using the outputs of translating the 4 images from a source class into all the remaining classes. Therefore, in our animal face dataset, total of 676 generated images are used for evaluation. 

In \textit{Landscapes} dataset, we manually classified the images into 7 different classes (beach, desert, forest, grass field, mountain, sea, snow). Each class has 300 different images except for desert class which have 100 different images. Since some classes have not enough number of images, we borrowed images from \textit{seasons} \citep{seasons} dataset. For metric calculation, we selected 8 testing images from each class, and used them as our content source images. Again, we translated the 8 images from source class into all the remaining classes. Therefore, a total of 336 generated images are used for our quantitative evaluation. 

For single image guided translation, we selected random images from AFHQ dataset for animal face translation; and for natural image generation, we selected random images from our \textit{Landscapes} datasets. 

\subsection{User Study Details}
For our user study in text-guided image translation task, we generated 130 different images using 13 different text conditions with our proposed and baseline models. Then we randomly selected 65 images and made 6 different questions. More specifically, we asked the participants question about three different parts: 1) Are the outputs have correct semantic of target text? (Text-match), 2) Are the generated images realistic? (Realism), 3) Do the outputs contain the content information of source images (Content). We randomly recruited a total of 30 users,  and provided them the questions using Google Form. The 30 different users come from age group 20s and 50s. We set the minimum score as 1, and the maximum score is 5. The users can score among 5 different options : 1-Very Unlikely, 2-Unlikely, 3-Normal, 4-Likely, 5-Very Likely. 

For the user study on image-guided translation task, we generated 40 different images using 8 different images conditions. Then we followed the same protocol to user study on text-guided image translation tasks, except for the content of questions. We asked the users in three different parts: 1) Are the outputs have correct semantic of target style image? (Style-match), 2) Are the generated images realistic? (Realism), 3) Do the outputs contain the content information of source images (Content).

% For our landscape dataset experiment,
% we translated 5 images from a source class into all the remaining classes,  total of 210 generated images are used for evaluation.

% For \textit{Landscapes} datasets, we manually filtered
\subsection{Algorithm}
For detailed explanation, we include Algorithm of our proposed image translation mathods in  Algorithm~\ref{alg:alg1}.
    \begin{algorithm}[t]
    \caption{Semantic image translation: given a diffusion score model $\epsilon_{\theta}(\x_t,t)$, CLIP model, and VIT model}
     \textbf{Input:} source image $\x_{src}$, diffusion steps $T$, resampling steps $N$, target text $\d_{trg}$, source text $\d_{src}$ or target image $\x_{trg}$\\
     \textbf{Output:} translated image $\hat{\x}$ which has semantic of $\d_{trg}$ (or $\x_{trg}$) and content of $\x_{src}$ \\
     $\x_T \sim \mathcal{N}(\sqrt{\Bar{\alpha}_t}\x_{src},(1-\Bar{\alpha}_t)\mathbf{I}) $, index for resampling $n = 0$
     % $x_{prev} \leftarrow x_k$
    \begin{algorithmic}[1]
        \ForAll{$t$ from $T$ to 0}
            % \If{$t$ is $k$}
            %     \For{$n$ from 0 to $N$}
            %         \State
            %     \EndFor
            % \EndIf
            \State $\epsilon \leftarrow \boldsymbol{\epsilon}_{\theta}(\x_t,t)$
            \State $\hat{\x}_0(\x_t) \leftarrow  \frac{\x_t}{\sqrt{\Bar{\alpha}_t}} - \frac{\sqrt{1-\Bar{\alpha}_t}}{\sqrt{\Bar{\alpha}_t}}\epsilon$
            \If{text-guided}
                \State $\nabla_{total} \leftarrow\nabla_{\x_t} \ell_{total}(\hat{\x}_0(\x_t);\d_{trg},\x_{src},\d_{src})$
            \ElsIf{image-guided}
                \State $\nabla_{total} \leftarrow\nabla_{\x_t} \ell_{total}(\hat{\x}_0(\x_t);\x_{trg},\x_{src})$
            \EndIf
            % \State $\x_{t-1} \sim \mathcal{N}(\mu+\Sigma\nabla_{total}, \Sigma)$
            \State $\z \sim \mathcal{N}(0,\mathbf{I})$
            \State $\x'_{t-1}=\frac{1}{\sqrt{\alpha_t}}\Big{(}\x_t-\frac{1-\alpha_t}{\sqrt{1-\bar{\alpha}_t}}\boldsymbol{\epsilon}\Big{)} +\sigma_t\z$
            \State $\x_{t-1}= \x'_{t-1} - \nabla_{total}$
            \If{$t$ = $T$ \textbf{and} $n < N$}
                \State $\x_t \leftarrow \mathcal{N}(\sqrt{1-\beta_{t-1}}\x_{t-1},\beta_{t-1}\mathbf{I})$
                \State $n \leftarrow n+1$
                \State \textbf{go to} 2
             \EndIf
        \EndFor 
        \State \textbf{return} $\x_{-1}$
    \end{algorithmic}
    \label{alg:alg1}
    \end{algorithm}

\section{Quantitative Ablation Study}
For more thorough evaluation of our proposed components, we report ablation study on quantitative metrics. In this experiment, we only used \textit{Animals} dataset due to the time limit. 
In Table \ref{table:abl}, we show the quantitative results on various settings. When we remove one of our acceleration strategies, in setting (b) and (f), we can see that the fid score is degraded as the outputs are not properly changed from the original source images. (e) When we use L2 maximization instead of our proposed $\ell_{sem}$, FID scores are improved from setting (b), but still the performance is not on par with our best settings. (d) When we use weak content regularization using LPIPS, we can see that the overall scores are degraded.   When we remove our proposed $\ell_{cont}$, we can observe that SFID and CSFID scores are lower than other settings. However, we can see that LPIPS score is severely high as the model hardly reflect the content information of original source images. (g) we use pre-trained VGG instead of using ViT for ablation study. Instead of ViT keys for structure loss, we substitute it with features extracted from VGG16 relu3\_1 activation layer. Also, we substitute ViT [CLS] token with VGG16 relu5\_1 feature as it contains high-level semantic features. We can see that the model shows decent SFID and CSFID scores, but the LPIPS score is very high. The result show that using VGG does not properly operate as regularization tool, rather it degrades the generation process with damaging the structural consistency. Overall, when using our best setting, we can obtain the best output considering all of the scores. 

\begin{table}[!t]

\begin{center}

\begin{tabular}{@{\extracolsep{5pt}}cccc@{}}
\hline
\multirow{2}{*}{\textbf{Settings}}  & \multicolumn{3}{c}{\textbf{Animals}}\\

\cline{2-4} 
 & SFID$\downarrow$& CSFID$\downarrow$&LPIPS$\downarrow$ \\
\hline
{VGG instead of ViT (g)} & {9.72}& {43.08} &	 {0.518}\\
No resampling (f) & 11.88&	59.09 &	 0.316\\
L2 Max instead of $\ell_{sem}$ (e) &13.18& 49.47 & 0.324 \\
% \cdashline{2-6}
LPIPs instead of $\ell_{cont}$ (d)& 11.15& 58.67& 0.400  \\
% \hline
No $\ell_{cont}$ (c)& 9.90&	33.07&	0.477  \\
No $\ell_{sem}$ (b) &15.00 &	53.43	& 0.347	\\
Ours (a) &9.98&	41.07	& 0.372	\\
\hline
\end{tabular}

% \end{adjustbox}
\end{center}
% }
% \vspace*{-.3cm}
\caption{Quantitative comparison of ablation studies. }
% \vspace{-0.3cm}
% \vspace*{-.3cm}
\label{table:abl}
% \end{threeparttable}
\end{table}

% \section{Quantitative Results on image-guided image translation}
% For evaluating our image-guided image translation method, we calculated single-pair quality measurement metric SIFID\cite{singan}. Since our baseline method of Splicing ViT\citep{splice} takes over 30 minutes for single image translation, we used 5 different image pairs to measure the metrics. For our style transfer baselines, SANet and WCT2 obtained 0.030 and 0.034, respectively. For our one-shot semantic transfer methods, STROTSS and Splicing Vit obtained 0.046 and 0.114, respectively. For Our method, we obtained 0.040 in SIFID.

% We can see that SANet and WCT2 scored lower than our method, as SIFID score is sensitive to color, not the perceptual semantics. 

\section{Artistic Style Transfer}
\begin{figure}[t!]
\centering
    \includegraphics[width=1.0\linewidth]{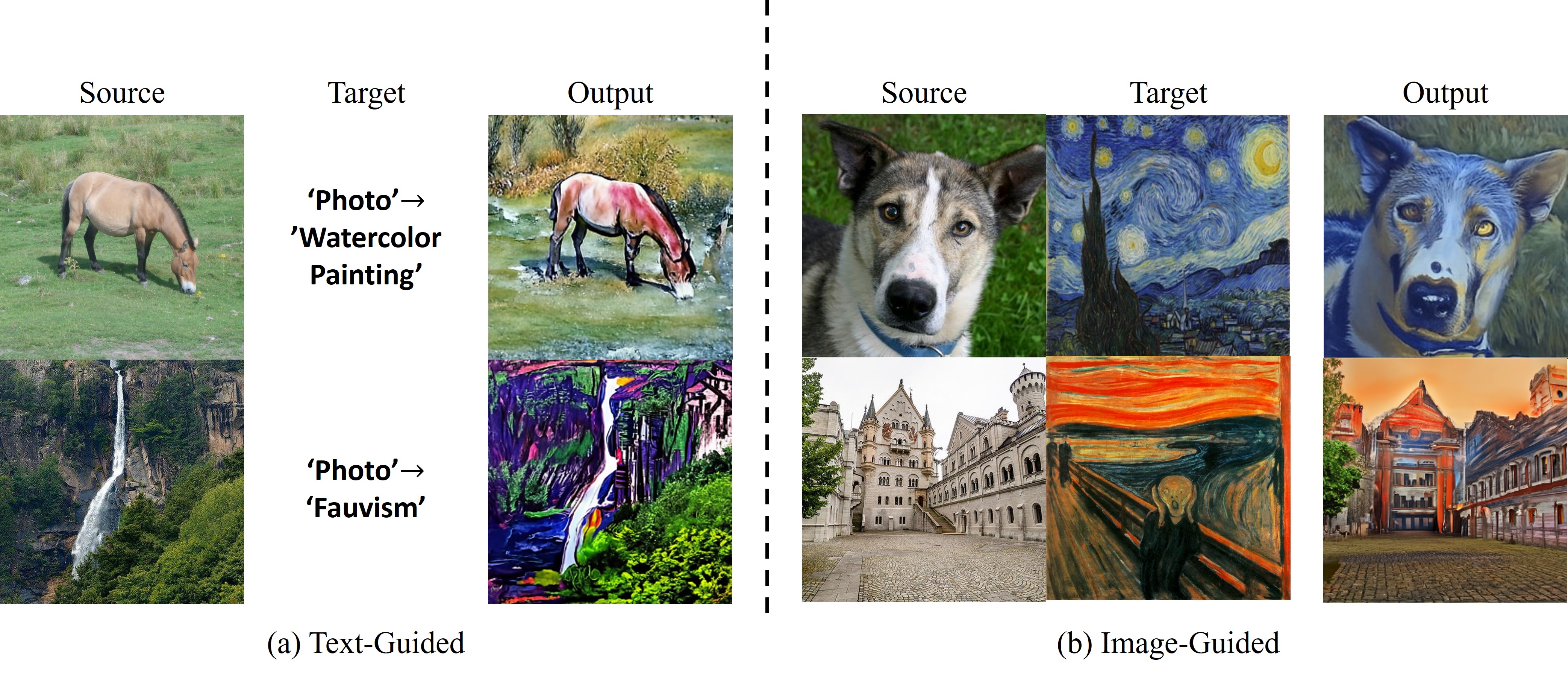}
    % \vspace{-0.3cm}
    \caption{Various outputs of artistic style transfer. We can translation natural images into artistic style paintings with both of text or image conditions.}
    \label{fig:fig_art}
    \end{figure}

With our framework, we can easily adapt our method to artistic style transfer. With simply changing the text conditions, or using artistic paintings as our image conditions, we can obtain the artistic style transfer results as shown in Fig.~\ref{fig:fig_art}. 

\section{Face image translation}

\begin{figure}[t!]
\centering
    \includegraphics[width=1.0\linewidth]{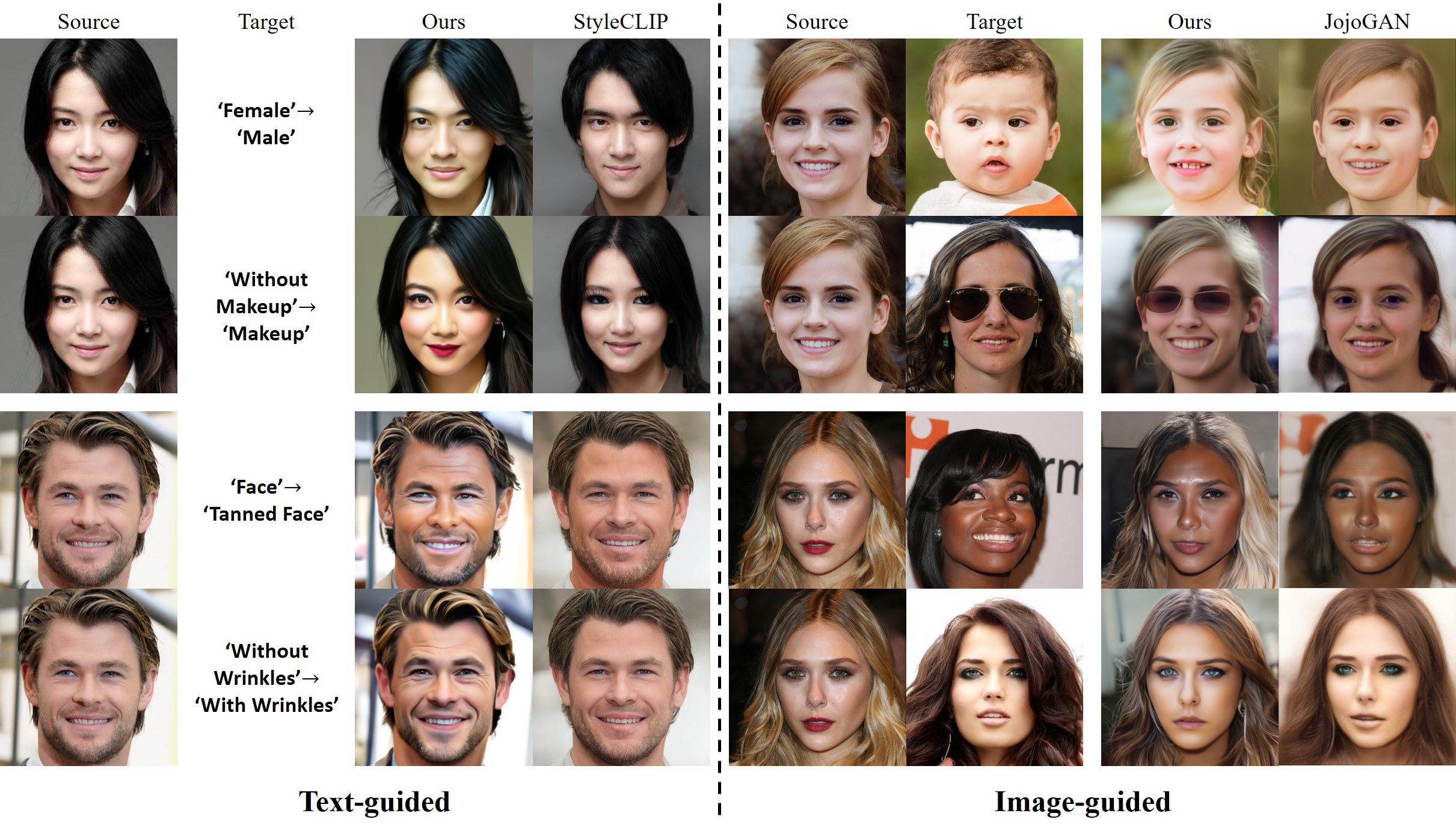}
    % \vspace{-0.3cm}
    \caption{Outputs from face image translation models. The outputs from our model successfully translated the human face images with proper target domain semantic information.}
    \label{fig:fig_face}
    \end{figure}

Instead of using score mode pre-trained on Imagenet dataset, we can use pre-trained score model on FFHQ human face dataset.  In order to keep the face identity between source and output images, we include $\lambda_{id}\ell_{id}$ which leverage pre-trained face identification model ArcFace\citep{arcface}. We calculate identity loss between $\x_{src}$ and denoised image $\hat{\x}_0(\x_t)$. We use $\lambda_{id}= 100$.

In Fig.~\ref{fig:fig_face}, we show that our method also can be used in face image translation tasks. For comparison, we included baseline models of face editing method StyleCLIP \citep{styleclip}, and one-shot face stylization model of JojoGAN \citep{jojogan}. The results show that our method can translate the source faces into target domain with proper semantic change. In baseline models, although some images show high quality outputs, in most cases the image failed in translating the images. Also, since the baseline models rely on pre-trained StyleGAN, they require additional GAN inversion process to translate the source image. Therefore, the content information is not perfectly matched to the source image due to the limitation of GAN inversion methods.

\section{Inference Time Comparison}

\begin{table}[!t]

\begin{center}

\begin{tabular}{@{\extracolsep{5pt}}c|ccccc@{}}
% \hline
 &\textbf{Ours}  &\textbf{SplicingVit} & \textbf{STROTSS} &\textbf{WCT2} & \textbf{SANET} \\
\hline
time & 37s &  25m 30s & 53s & 0.18s & 0.12s	\\
% \hline
\end{tabular}

% \end{adjustbox}
\end{center}
% }
% \vspace*{-.3cm}
\caption{Quantitative comparison on inference times of image-guided translation models. }
% \vspace{-0.3cm}
% \vspace*{-.3cm}
\label{table:time}
% \end{threeparttable}
\end{table}

{To evaluate the time-efficiency of our method, we calculate the inference times of the various image-guided translation models. All experiments are conducted with single RTX3090 GPU, on the same hardware and software environment. We use the images of resolution 256$\times$256 for experiments. In Table \ref{table:time}, we compare the times taken for single image translation. For single-shot semantic transfer models of Splicing ViT, the inference time is relatively long as we need to optimize large U-Net model for each image translation. In STROTSS, it requires texture matching calculation for single image translation, so it takes long time. For arbitrary style transfer models of WCT2 and SANet, the inference is done with only single-step network forward process, as the model is already trained with large dataset. Our model takes about 40 seconds, which is moderate when compared to the one-shot semantic transfer models (SplicingVit,STROTSS). However, the time is still longer than the style transfer models, as our model need multiple reverse DDPM steps for inference. In the future work, we are planning to improve the inference time with leveraging recent approaches. }

\section{Semantic Segmentation Outputs}
\begin{figure}[t!]
\centering
    \includegraphics[width=1.0\linewidth]{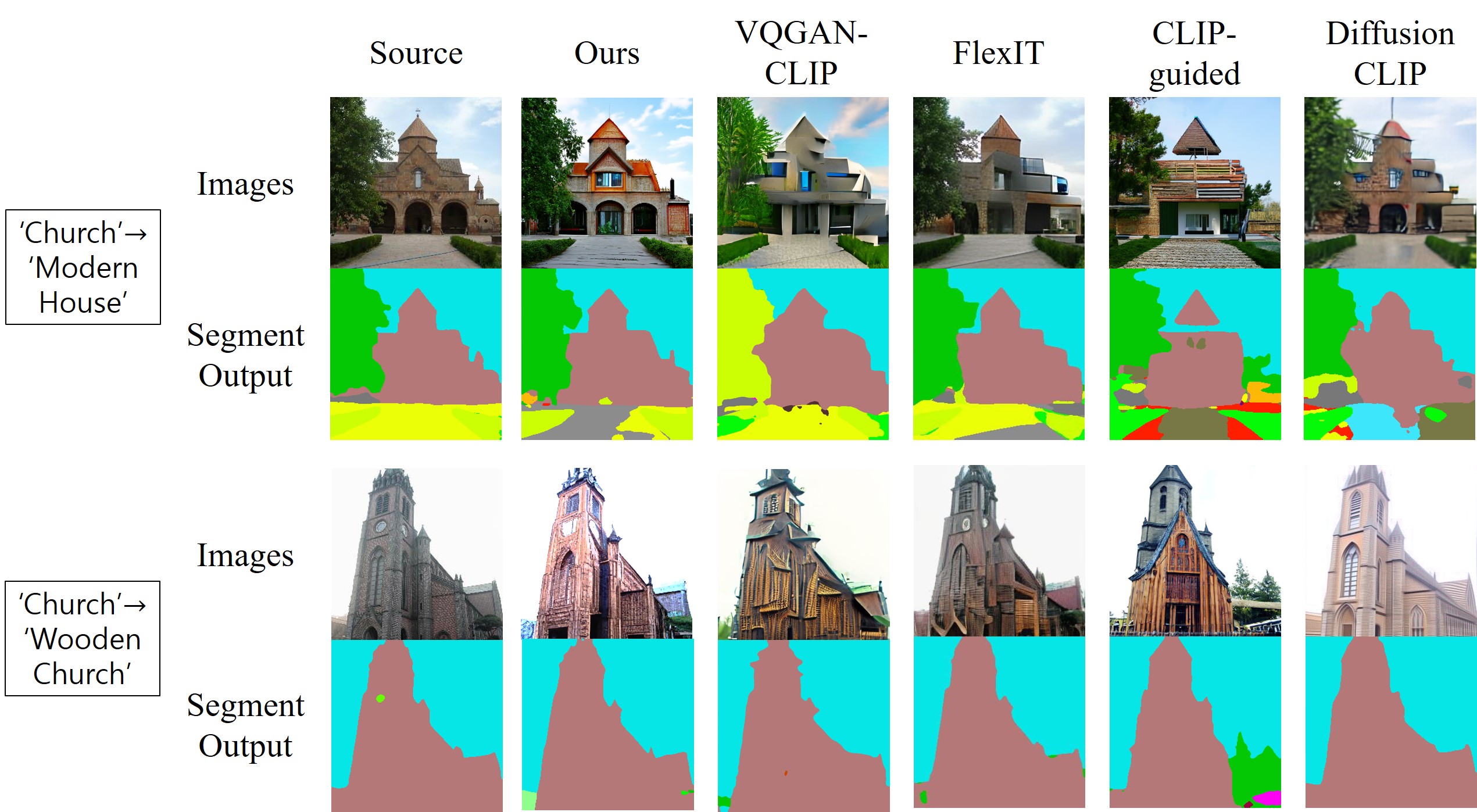}
    % \vspace{-0.3cm}
    \caption{{Comparison results on semantic segmentation maps from baseline outputs. When comparing segmentation maps, our model outputs show high structural consistency with the source images.}}
    \label{fig:fig_seg}
    \end{figure}

{To further verify the structural consistency between output and source images, we compared the semantic segmentation maps from outputs and source images. For experiment, we use semantic segmentation model \citep{segmentation} which is pre-trained on ADE20K dataset. We referenced the official source code\footnote{\url{https://github.com/CSAILVision/semantic-segmentation-pytorch}} for segmentation model. Figure~\ref{fig:fig_seg} shows the comparison results. In case of the baseline models VQGAN-CLIP, CLIP-guided diffusion, we can see that the segmentation maps are not properly aligned to the source maps, which means the model cannot keep the structure of source images. In case of FlexIT, the model outputs maps have high similarity to the source maps, but the semantic change is not properly applied. In our model, we can see the output maps have high similarity to the source maps, while semantic information is properly changed. }

\section{Additional Comparison on Image-guided Translation}
\begin{figure}[t!]
\centering
    \includegraphics[width=0.7\linewidth]{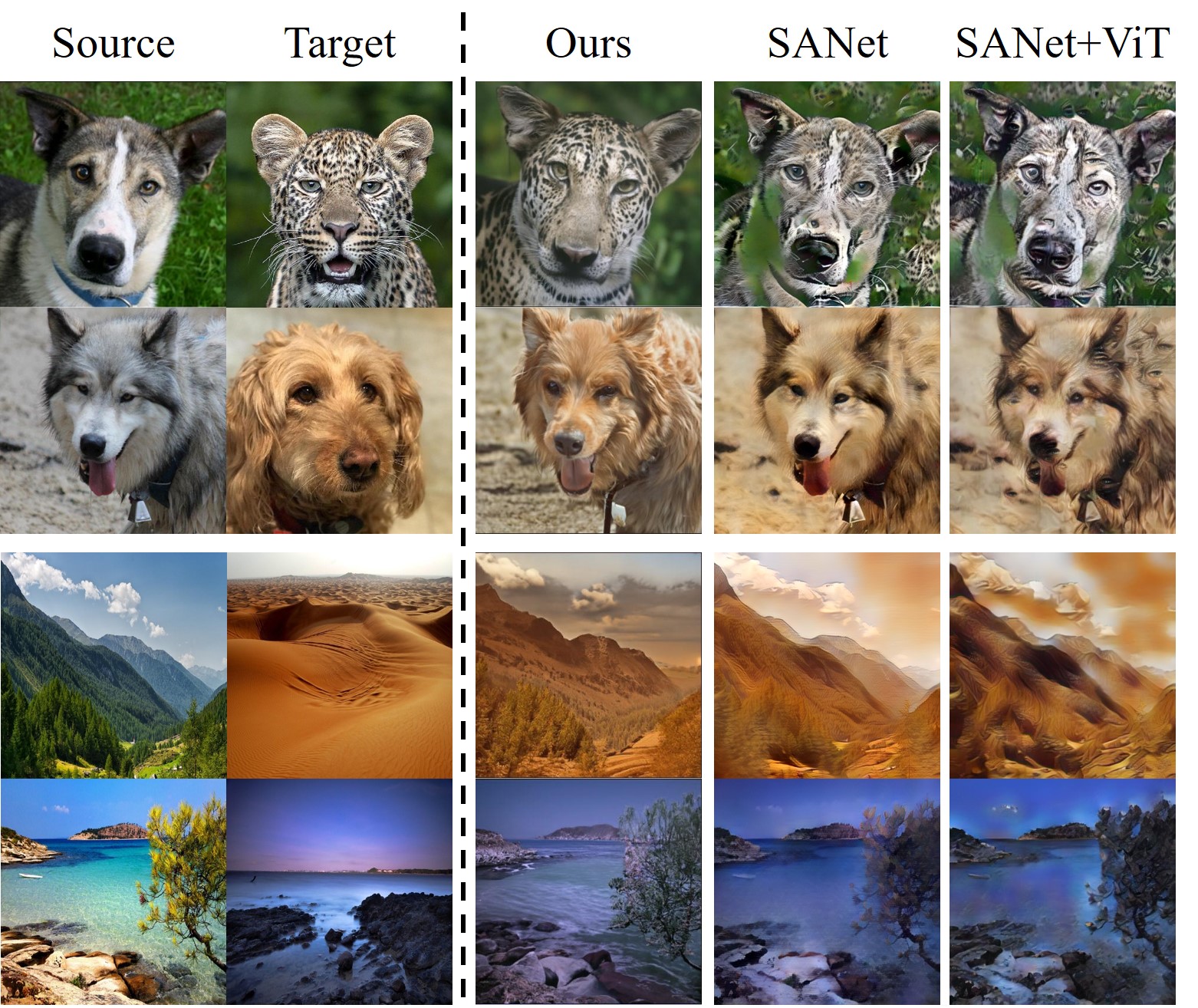}
    % \vspace{-0.3cm}
    \caption{Additional comparison on image-guided translation. For fair experiment conditioning, we trained the baseline SANet with ViT-based losses.}
    \label{fig:fig_sanet}
    \end{figure}

\begin{figure}[t!]
\centering
    \includegraphics[width=0.6\linewidth]{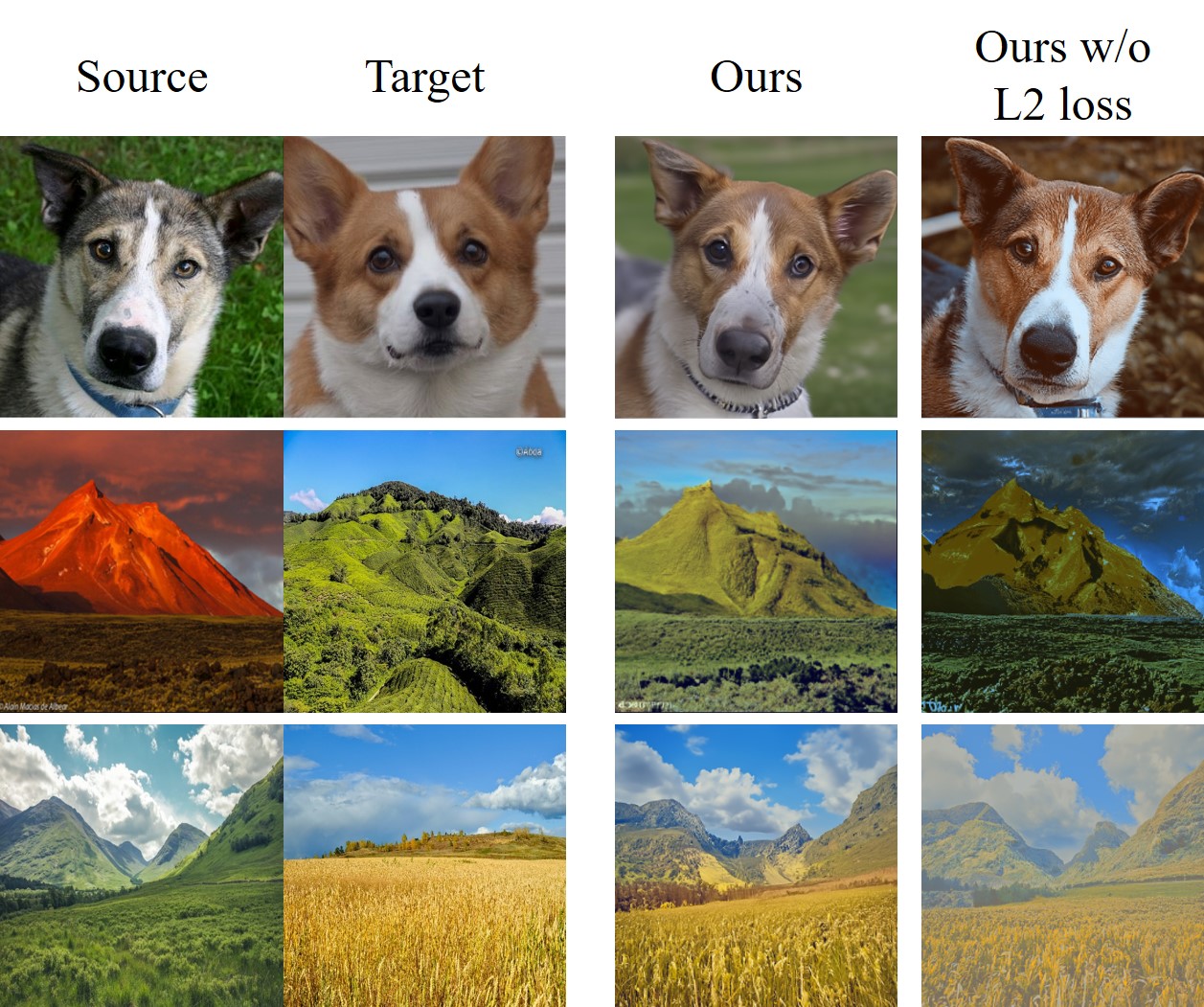}
    % \vspace{-0.3cm}
    \caption{Ablation study results on pixel-wise l2 loss. Without pixel loss, the output image color is not matched to the color scale of the target images.}
    \label{fig:fig_l2}
    \end{figure}
    
 For fair comparison with the baseline models, we trained the baseline SANet with our proposed ViT-based loss functions. When we train SANet with replacing the existing style and content loss with our $\ell_{cont}$ and $\ell_{sty}$, we found that the training is not properly working. Therefore, we simultaneously used existing VGG-based style and content loss with our proposed ViT-based losses. In Fig.~\ref{fig:fig_sanet}, we can see that when training SANet with ViT, the results still show incomplete semantic transfer results. Although the output seems to contain more complex textures than basic model, the model performance is still confined to simple color transformation.  

 To further evaluate the effect of pixel-wixe l2 loss for image-guided translation, we conducted additional experiments in Fig. ~\ref{fig:fig_l2}. When we remove the pixel-wise l2 loss in our image-guided translation task, we can see the semantic of output images follow the target images, but the overall color of the output images are slightly unaligned with the target image color.
The result show that using weak l2 loss help the model to accurately apply the color of target images to outputs.

\section{Limitation and Future work}

\begin{figure}[t!]
\centering
    \includegraphics[width=0.9\linewidth]{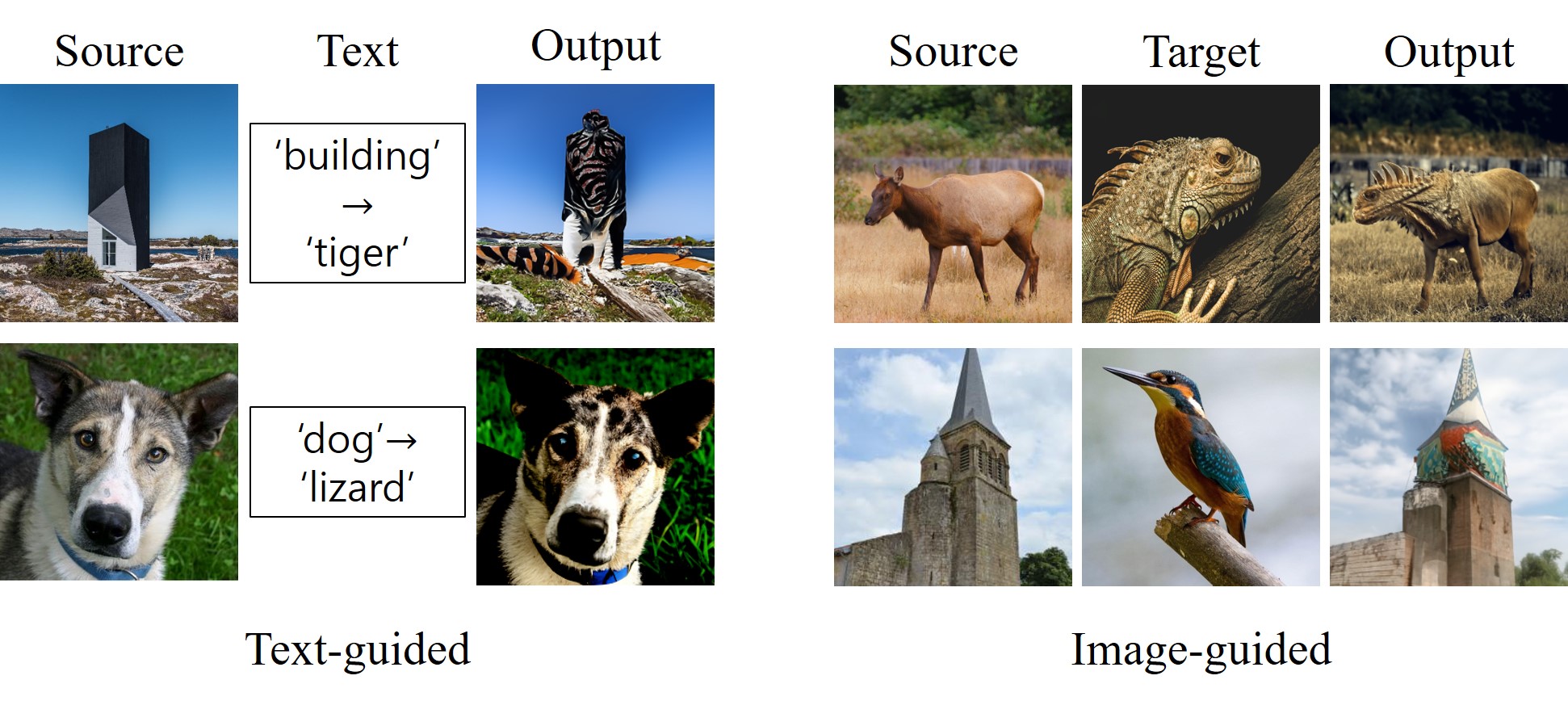}
    % \vspace{-0.3cm}
    \caption{{Failure case outputs. If the semantic distance between source and target conditions are extremely far, semantic translation sometimes fails.}}
    \label{fig:fig_failure}
    \end{figure}
    
Although our method has shown successful performance in image conversion, it still has limitations to solve.  First, if the semantic distance between the source image and the target domain is too far (e.g building $\rightarrow$ Tiger), the output is not translated properly as shown in Fig.~\ref{fig:fig_failure}. We conjecture that this occurs when the text-image embedding space in the CLIP model is not accurately aligned, therefore it can be solved by using the advanced text-to-image embedding model. Second, our method has limitation that the image generation quality heavily relies on the performance of the pre-trained score model. This can also be solved if we use a diffusion model backbone with better performance. In future work, we plan to improve our proposed method in these two directions.

\section{Additional Results}
\begin{figure}[t!]
\centering
    \includegraphics[width=1.0\linewidth]{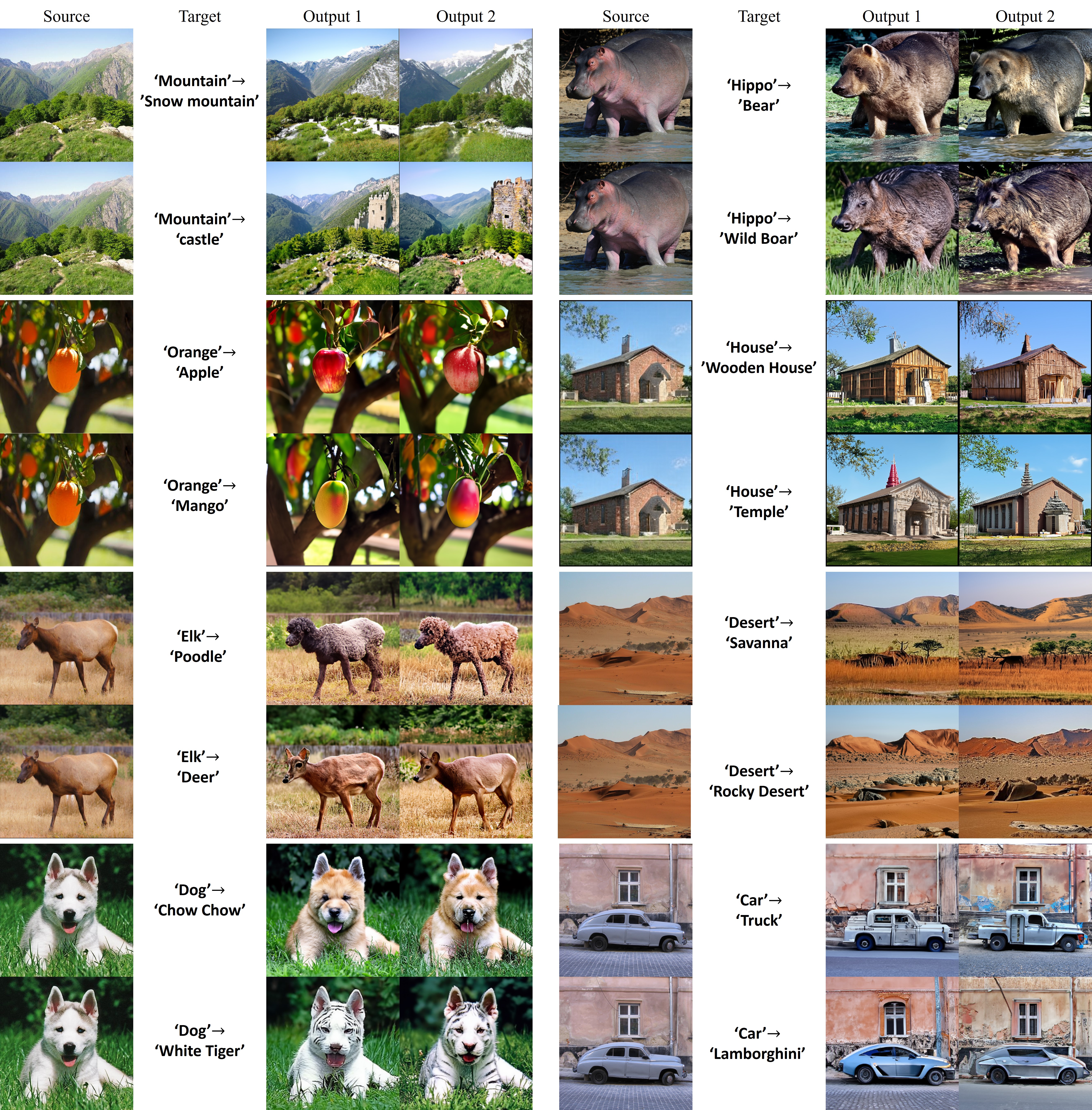}
    % \vspace{-0.3cm}
    \caption{Qualitative results of text-guided image translation.}
    \label{fig:fig_text_add}
    \end{figure}

\begin{figure}[t!]
\centering
    \includegraphics[width=1.0\linewidth]{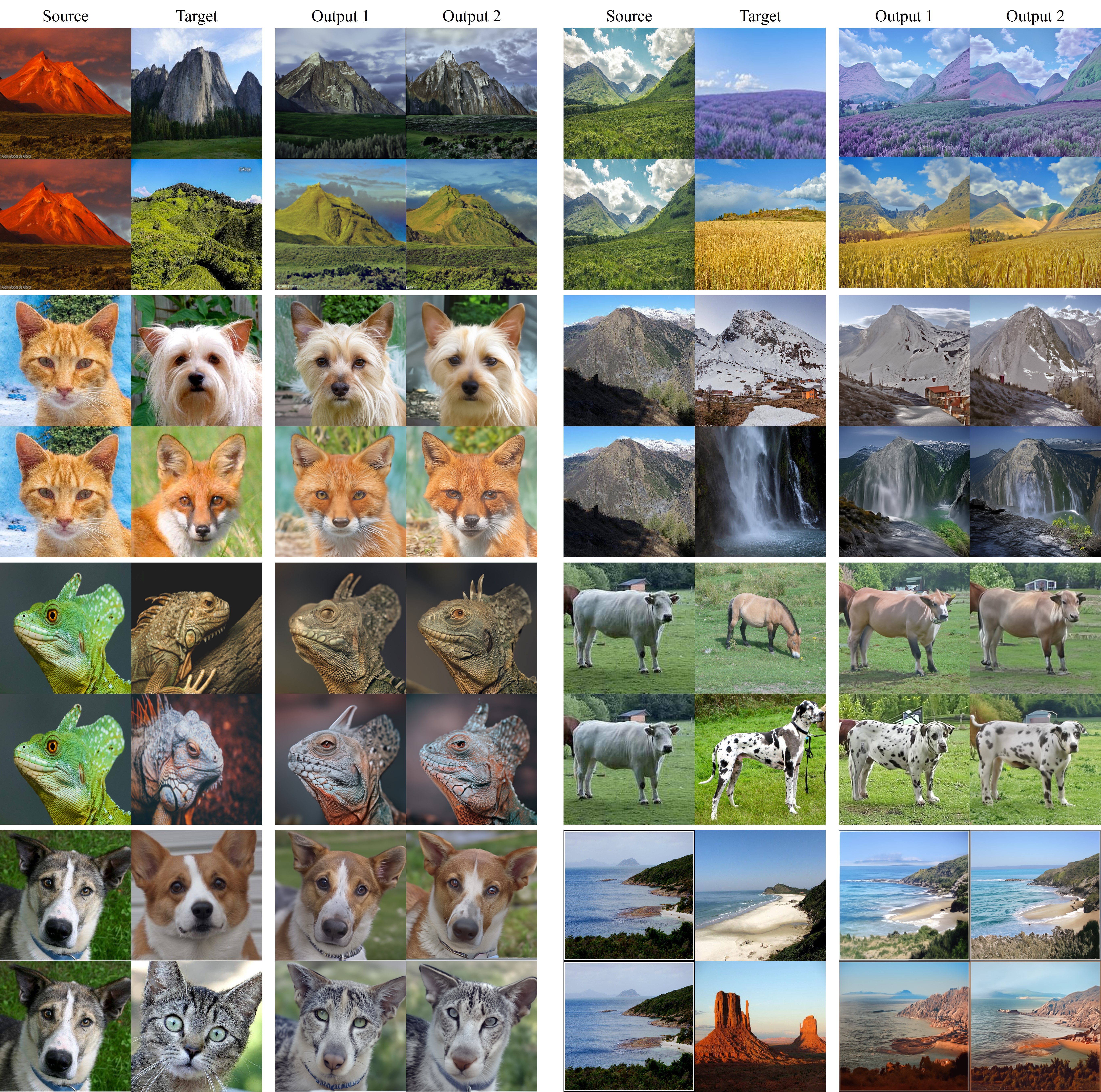}
    % \vspace{-0.3cm}
    \caption{Qualitative results of image-guided image translation.}
    \label{fig:fig_img_add}
    \end{figure}

For additional results , in Fig. \ref{fig:fig_text_add} we show the image translation outputs using text conditions. 
In Fig. \ref{fig:fig_img_add}, we additionally show the results from our image-guided image translation. 
We can successfully change the semantic of various natural images with text and image conditions.

\newpage

\newpage

\end{document}